\documentclass[10pt,twocolumn,letterpaper]{article}

\usepackage[pagenumbers]{cvpr}  

\usepackage[nopar]{lipsum}
\usepackage{graphicx}
\usepackage{capt-of}
\usepackage{xcolor}
\usepackage{overpic}
\usepackage{amsmath,amssymb,amsfonts,dsfont,pifont,bm,bbm,mathrsfs,mathtools,nicefrac}
\usepackage{algorithm,algpseudocode,listings}
\usepackage{booktabs,multirow,adjustbox,diagbox,threeparttable}
\definecolor{citeblue}{RGB}{48,111,186}
\usepackage[pagebackref,breaklinks,colorlinks,citecolor=citeblue,bookmarks=false]{hyperref}
\usepackage[capitalize]{cleveref}  
\crefname{section}{Sec.}{Secs.}
\Crefname{section}{Section}{Sections}
\crefname{table}{Tab.}{Tabs.}
\Crefname{table}{Table}{Tables}
\crefname{figure}{Fig.}{Figs.}
\Crefname{figure}{Figure}{Figures}
\crefname{equation}{Eq.}{Eqs.}
\Crefname{equation}{Equation}{Equations}
\newcommand{\tocite}[1]{\textcolor{red}{[TO CITE]}}

\newcommand\blfootnote[1]{%
  \begingroup
  \renewcommand\thefootnote{}\footnote{#1}%
  \addtocounter{footnote}{-1}%
  \endgroup
}

\begin{document}

\title{Uncertainty-Aware Optimal Transport for Semantically Coherent Out-of-Distribution Detection}

\author{Fan Lu$^{1,*}$\qquad~Kai Zhu$^{1,*,\ddagger}$\qquad~Wei Zhai$^{1}$\qquad~Kecheng Zheng$^{2}$\qquad~Yang Cao$^{1,3,\dagger}$\\
{$^{1}$~University of Science and Technology of China} \qquad
{$^{2}$~Ant Group}\\
{$^{3}$~Institute of Artificial Intelligence, Hefei Comprehensive National Science Center}\\
\small{\texttt{\{lufan@mail., zkzy@mail., wzhai056@\}ustc.edu.cn}} \qquad
\small{\texttt{zkccloud@gmail.com}} \qquad
\small{\texttt{forrest@ustc.edu.cn}}
  }
\maketitle

\blfootnote{$*$Co-first Author. $\dagger$Corresponding Author. $\ddagger$Work done during an internship at Ant Group.}

\begin{abstract}

Semantically coherent out-of-distribution (SCOOD) detection aims to discern outliers from the intended data distribution with access to unlabeled extra set. The coexistence of in-distribution and out-of-distribution samples will exacerbate the model overfitting when no distinction is made. To address this problem, we propose a novel uncertainty-aware optimal transport scheme. Our scheme consists of an energy-based transport (ET) mechanism that estimates the fluctuating cost of uncertainty to promote the assignment of semantic-agnostic representation, and an inter-cluster extension strategy that enhances the discrimination of semantic property among different clusters by widening the corresponding margin distance. Furthermore, a T-energy score is presented to mitigate the magnitude gap between the parallel transport and classifier branches. Extensive experiments on two standard SCOOD benchmarks demonstrate the above-par OOD detection performance, outperforming the state-of-the-art methods by a margin of $27.69\%$ and $34.4\%$ on FPR@95, respectively. Code is available at \href{https://github.com/LuFan31/ET-OOD}{https://github.com/LuFan31/ET-OOD}.

\end{abstract}

\section{Introduction}\label{sec:intro}

\begin{figure}[t]
\centering
\begin{overpic}[width=0.99\linewidth]{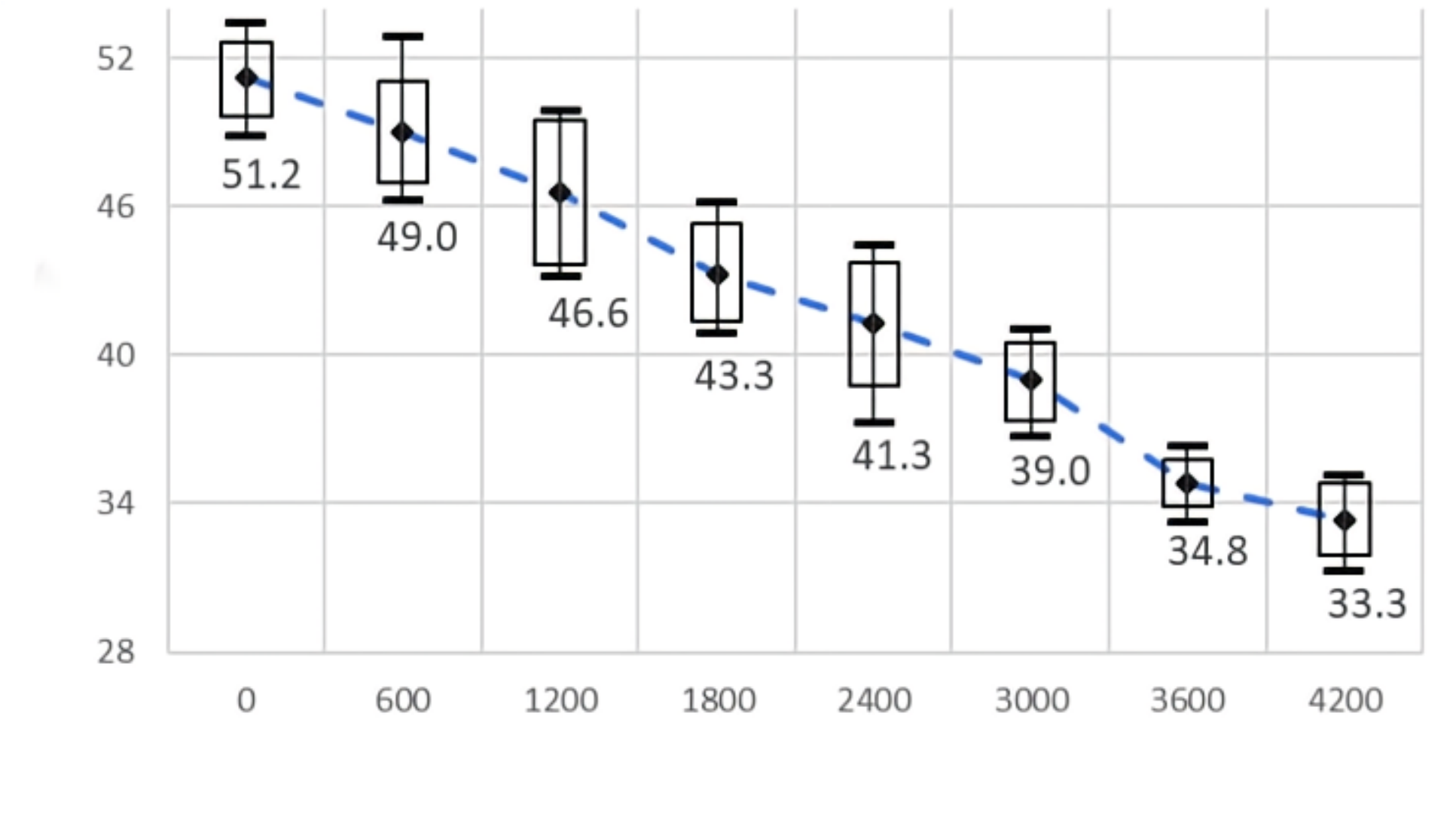}
  \put(27, 0.2){\textbf{\footnotesize{Number of Accurately Assigned Labels}}}
  \put(0.3, 22){\rotatebox{90}{\footnotesize\textbf{FPR@95 (\%)}}}
\end{overpic}
\vspace{-0.2em}
\caption{\textbf{The effect of the number of accurately assigned labels.} We gradually increase the ground-truth labels of unlabeled ID samples during the baseline~\cite{hendrycks18oe} training and report the FPR@95$\%$ (lower value is better) results averaged by multiple tests. The trend shows that more accurately assigned labels
enhance the OOD detection capability of the model.}
\vspace{-10pt}
\label{fig:Fig1}
\end{figure}

Deep learning models trained in close-set world often suffer from performance degradation in real-world scenarios due to the interference of out-of-distribution (OOD) inputs---unknown samples whose classes don't overlap with those already seen during training~\cite{zhou2022domain}. Full-supervised models tend to make overconfident predictions on OOD inputs~\cite{nguyen2015deep,scheirer2012toward}
, which severely limits the application in high-risk fields such as autonomous driving~\cite{autodriving} and medical analysis~\cite{schlegl2017unsupervised}. To this end, OOD detection~\cite{baseline,odin,mcd,energyood,yang2021semantically} aims to obtain a reliable model to identify these unknown samples as abnormal ones and reject them at test time. 

\begin{figure*}[t]
  \centering
  \includegraphics[width=0.90\linewidth]{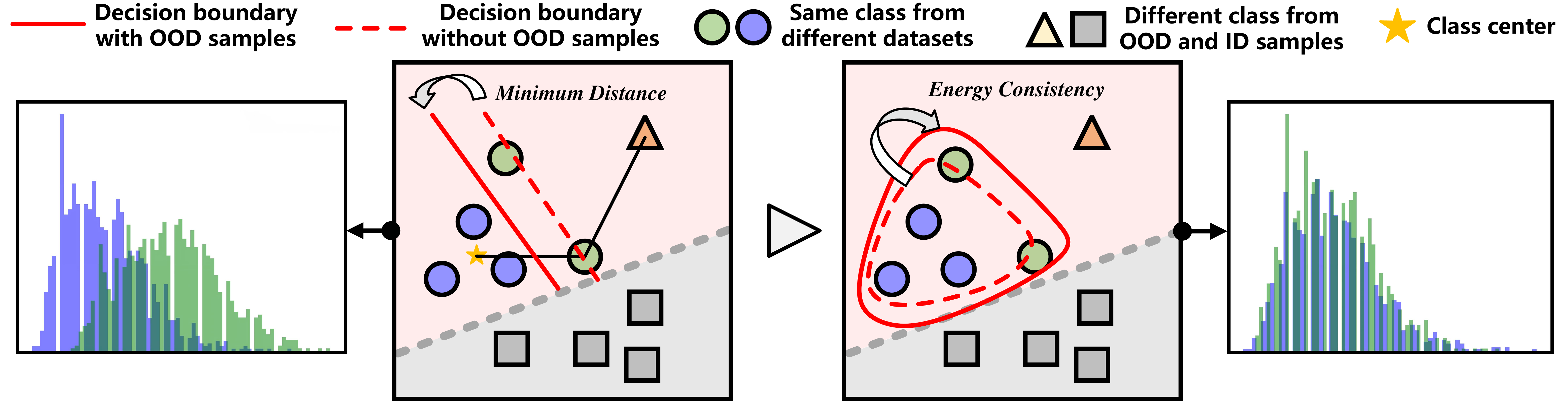}
  \vspace{-8pt}
  \caption{\textbf{Motivation of the proposed method}. When an extra unlabeled dataset is introduced into training, the resulting covariate shifts will easily confuse the minimum feature distance (\textit{e.g., Euclidean distance}) from the correct class mean with other wrong classes. The left histogram reflects the minimum Euclidean distance distribution from the class mean for the `dog' images in both CIFAR-10 (blue) and tiny-ImageNet (green). Unfortunately, although belonging to the same class, their feature distance distributions differ greatly, which limits the ability of cluster-based methods to mine unlabeled ID semantic knowledge in the SCOOD setting. In contrast, as shown in the right histogram, the distributions of the `dog' images in the two different datasets maintain good consistency on the energy metric, facilitating a more stable semantic assignment process.}
  \label{fig:motivation}
  \vspace{-5pt}
\end{figure*}

To learn the discrepancy between ID/OOD samples better, a rich line of OOD detection methods like OE~\cite{hendrycks18oe} and MCD~\cite{mcd} utilizing extra OOD data during training has been developed. OE flattens the prediction probability of OOD samples among all ID classes, and MCD maximizes the entropy discrepancy between ID and OOD samples outputted by two parallel classifiers. Although extra data greatly enhances the performance of models across traditional OOD benchmarks, these benchmarks still face two problems: \textbf{1)} to ensure the sampling effectiveness,  ID samples in extra unlabeled datasets are manually removed to obtain the pure OOD set, which does not correspond to reality since ID samples are inevitably mixed into the unlabeled set and expensive to be purified. \textbf{2)} the great improvement on them is most likely due to the overfitting on the training OOD samples~\cite{yang2021semantically}, so the model paying attention to the lower-level covariate shifts between datasets has difficulty generalizing to other OOD domains, especially when the scale of the dataset is not large.

To overcome these challenges, the SCOOD benchmarks \cite{yang2021semantically} are proposed to urge OOD detection models to focus on distinguishing semantic discrepancy between ID/OOD samples and avoid overfitting low-level covariate shifts on different data sources. K-means clustering is adopted to dynamically filter ID samples from the unlabeled set during training, enhancing the optimization for ID classification~\cite{yang2021semantically}. Unfortunately, we statistically observe that this strategy can only assign correct semantic labels to less than 1$\%$ unlabeled ID samples at each epoch, which means a large number of samples are added to the ID classification training with wrong labels and undoubtedly limits the ability of the model to fully understand the semantic discrepancy between ID/OOD samples.


To investigate the impact of accurately assigned semantic labels on SCOOD tasks, we gradually increase the ground-truth labels of unlabeled ID samples used in training with the baseline in \cite{yang2021semantically}, and then observe the variation trend of model performance shown in~\cref{fig:Fig1}. The model performance consistently improves as the correctly assigned labels increase, leading the key question of how to allocate as many ID-related samples in the unlabeled set correctly as possible. Our further analysis indicates that compared to the traditional feature distance (\textit{e.g., Euclidean distance}), the class-specific energy score remains consistent even when the assigned cluster contains shifted unlabeled samples with an ID class. As shown in \cref{fig:motivation}, due to the covariate shifts of the same class samples from another dataset with a black circle, the minimum distance with the correct class mean is easily confused with other clusters. In contrast, the energy-based uncertainty prior can effectively mitigate the corresponding interference owing to the intra-class interaction.

To effectively involve energy-based guidance in the assignment process, we propose an uncertainty-aware optimal transport scheme. This core energy-based transport (ET) mechanism estimates the energy transport cost to encourage lower-cost samples to be assigned to the same cluster and the higher-cost ones to be split uniformly among all clusters, leading to a semantic-consistent distribution. Specifically, the energy score is obtained by performing the aggregation operation in logit space.

To further enhance the assignment efficiency of the aforementioned transport mechanism, an inter-cluster extension strategy is proposed to narrow the margin of the same ID class while widening the margin between ID and OOD samples. An enhanced global feature representation will be mapped to the lower-dimensional logit space, from where we can obtain a more discriminative prediction distribution for each cluster and class-specific energy score better reflecting the semantic information of samples. By utilizing both of them, the semantics among samples assigned to different clusters by ET will be more discriminative.
During inference, we also choose the energy score with temperature scaling produced by the classifier as the detection metric, aligning with the transport optimization more closely.

Our contributions are summarized as follows: \textbf{1)} We analyze the limitations of the clustering strategy for the SCOOD task and further explore how to address the fundamental challenge of OOD detection. \textbf{2)} We propose a novel uncertainty-aware optimal transport scheme to fully utilize the semantic consistency hidden in the unlabeled set, resulting in the covariate-invariant assignment. \textbf{3)} Extensive experiments demonstrate that the proposed method achieves state-of-the-art performance on SCOOD benchmarks.

\section{Related Work}\label{sec:related-work}
\subsection{OOD Detection with Extra Data}
The mainstream OOD detection works primarily focus on detecting samples with semantic shifts, \textit{i.e.}OOD samples drawn from classes that do not overlap with known classes. To enhance the ability to understand ID/OOD discrepancy, some methods synthesize OOD training data such as \cite{lee2018training,sricharan2018building,vernekar2019out,duvos}, and another series of OOD detection methods introduce extra OOD data into training. OE~\cite{hendrycks18oe} uses purified unlabeled OOD data, which is encouraged to produce a uniform softmax distribution during training. MCD~\cite{mcd} maximizes the entropy discrepancy between ID and extra OOD samples outputted by two parallel classifiers. The survey~\cite{yang2021generalized} indicates that OOD detection with external data generally performs better. However, massive unlabeled samples could be very easily obtained considering their accessibility, and ID data would inevitably exist among them. Hence, `purified unlabeled OOD data' means the mixed ID data need to be manually removed, which brings a high cost and is impractical. 

Against the above issues, UDG~\cite{yang2021semantically} proposes a realistic benchmark that retains ID samples in the unlabeled set, namely SCOOD (Semantically Coherent Out-of-Distribution Detection) benchmarks. UDG separates the ID/OOD samples in the unlabeled dataset based on K-means clustering, and adds the ID samples to the labeled samples according to semantics for training together. During training, UDG focuses on the semantic shifts between samples, which is a fundamental problem to be solved in OOD detection tasks, and well explores the semantic knowledge in the unlabeled set. UDG underlines that the OOD detection task should concentrate on semantic shifts, but in experiments we found that the clustering method cannot sufficiently collect unlabeled ID samples and can only assign less than 1$\%$ correct semantic labels to unlabeled ID samples at each epoch. In this paper, we propose an uncertainty-aware optimal transport scheme to achieve more accurate label assignment and learn a more significant difference between ID and OOD semantics.

\subsection{Optimal Transport}
wGAN~\cite{wGAN} applies optimal transport (OT) theory to the field of computer vision and has received extensive attention. Its basic idea is to minimize the Wasserstein distance between the distribution of sampled data and the image distribution synthesized by deep generative models. In summary, OT is to transform one distribution into another with minimal cost (\textit{e.g.,} Wasserstein distance), so it is applied to distribution matching tasks such as few-shot classification~\cite{guo2022adaptive} and domain adaptation~\cite{turrisi2022multi,li2020mutual}. In close-set unsupervised classification tasks, OT can assign pseudo-labels for unlabeled samples. SeLa~\cite{asano2019self} implements equipartition constraint on the label assignment matrix $
\mathbf{Q}$, and introduces KL divergence as a regularization term into the optimization objective, then uses a fast version of the Sinkhorn-Knopp~\cite{cuturi2013sinkhorn} algorithm to assign labels. SwAV~\cite{caron2020unsupervised} keeps the soft-label assignment produced by the Sinkhorn-Knopp algorithm without approximating it into a one-hot assignment like SeLa. For the SCOOD task, we adopt an individual OT strategy called energy-based transport (ET) mechanism, which introduces the energy score as the transport cost to guide the cluster distribution of samples.
\section{Method}\label{sec:method}

\subsection{Problem Setting}

In this section, we will introduce the setting of SCOOD task. Firstly, we denote in-distribution and out-of-distribution set as $I$ and $O$, respectively. The training set $D$ on SCOOD benchmarks consists of labeled training set $D_L$ and unlabeled training set $D_U$. While the samples contained in $D_L$ are all from $I$, the $D_U$ is a mixture of partial ID set $D^I_U$ from $I$ and OOD set $D^O_U$ from $O$, which is unlike previous OOD benchmarks. So the training set can be represented as $D=(D_L\subset I)\cup(D^I_U\subset I)\cup(D^O_U\subset O)$. Likewise, the test set $T$ comprises $T^I\subset I$ and $T^O\subset O$. We desire to assign correct ID labels to as many samples in $D^I_U$ as possible during training, and identify data from $T^O$ as negative (OOD) samples and classify samples from $T^I$ correctly at test time.

\subsection{Overall Pipeline}  
\begin{figure}[t]
  \centering
\includegraphics[width=0.98\linewidth]{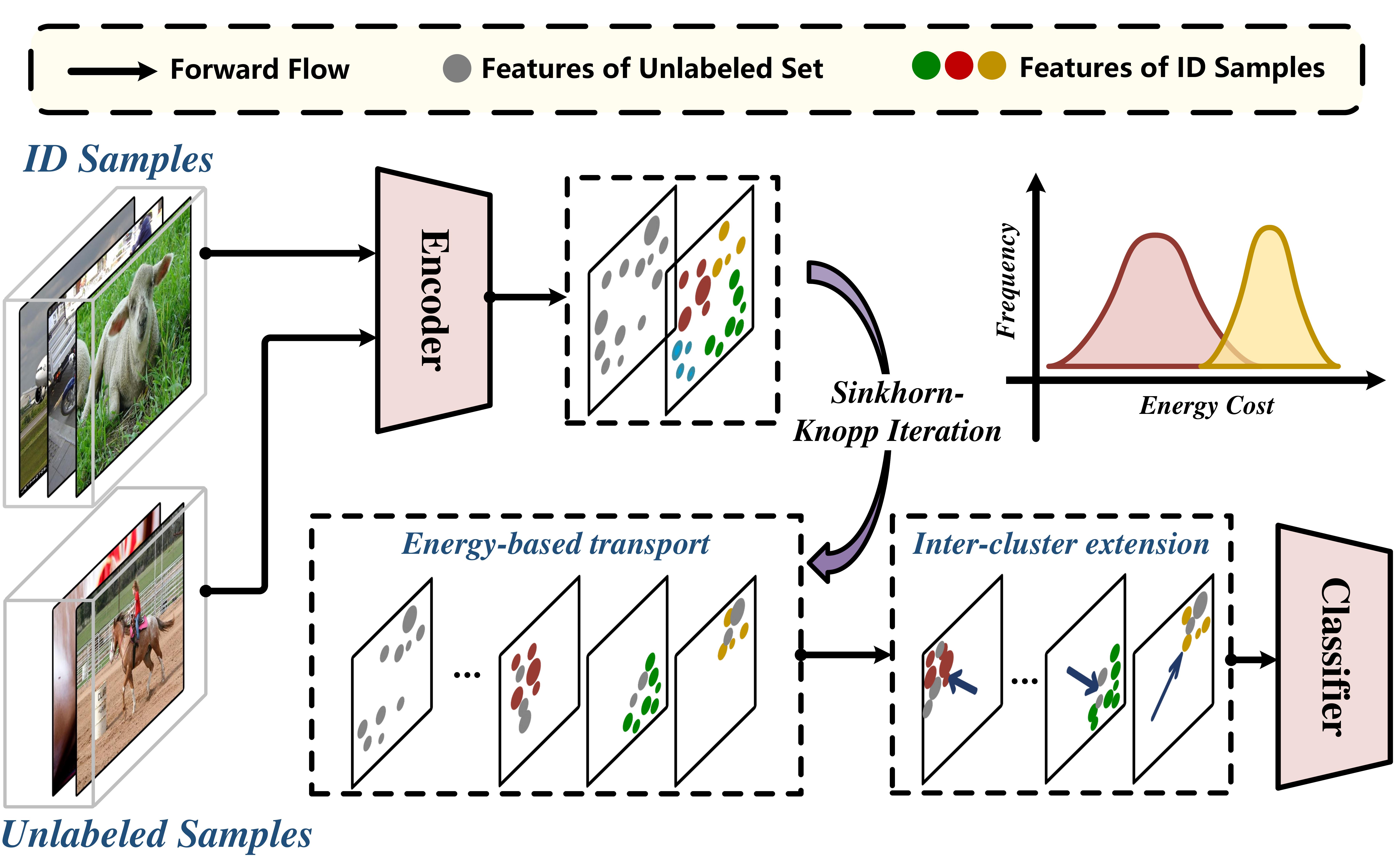}
  \caption{\textbf{Overall framework of our uncertainty-aware optimal transport scheme.} The energy scores for both ID and OOD samples are firstly calculated from output logit and then utilized to estimate the energy-based transport cost, guiding the assignment of unlabeled ID samples during the Sinkhorn-Knopp iteration. Finally, the energy scores will be further enhanced by the unlabeled samples obtaining correct ID labels and the inter-cluster extension strategy, facilitating subsequent assignment and classification.}
  \label{fig:model}
  \vspace{-0.28em}
\end{figure}

 As shown in Fig. \ref{fig:model}, we design a novel uncertainty-aware optimal transport (OT) scheme, which consists of an energy-based transport (ET) mechanism and an inter-cluster extension strategy $ L_{rep}$ to assign correct ID class label to partial unlabeled ID samples and then involve them into the joint training. The ET introduces the energy score as uncertainty and estimates the uncertainty-based transport cost to guide the cluster distribution of all samples. To further promote the discrimination in logit space, from which the uncertainty can significantly reflect the difference between ID/OOD samples, the inter-cluster extension strategy enhances the global feature representation mixed with ID and OOD samples and then the enhanced representation will be mapped into more discriminate logits.
 
\subsection{Energy-Based Transport Mechanism}
Unlike assigning labels to unlabeled samples in the close-set classification tasks, the situation we face is more challenging because we have to overcome the interference of OOD samples when assigning labels. To address it, we propose an energy-based transport (ET) mechanism based on logit space to explore the semantic discrepancy and knowledge hidden in the unlabeled set more sufficiently. 

Specifically, we add a $K$-dimensional OT head $h_{ot}$ in parallel with the classification head $h_{cls}$ to map training samples to $K$ clusters. Given the set of all $N$ samples $\mathcal{X}= \{x_1,x_2,\dots,x_N\}$ and the set of their feature extracted by encoder $\mathcal{Z}= \{z_1,z_2,\dots,z_N\}$, we denote the probability of the $i\mbox{-}th$ sample over all clusters as $p(\mathbf{c}|x_i)=softmax(h_{ot}(z_i))$, where $\mathbf{c}$ is a $K$-dimension vector representing $K$ clusters. Then we can define the cost matrix in optimal transport problem as $\mathbf{P}\in\mathbb{R}^{K\times N}$ and $P_{ji}=p(c_j|x_i)$ denotes the probability of $x_i$ belonging to $c_j$, where $c_j$ is the $j\mbox{-}th$ cluster. Likewise, we represent $\mathbf{Q}\in\mathbb{R}^{K\times N}$ as the assignment matrix and $Q_{ji}=q(c_j|x_i)$ denotes the posterior probability of $x_i$ assigned to $c_j$. Note that the assignment matrix $\mathbf{Q}$ just represents the assignment of a certain cluster to each sample, rather than directly assigning labels for classification training. When we represent the distributions of $N$ samples and $K$ clusters using an $N$-dimension vector $\boldsymbol{\beta}$ and a $K$-dimensional vector $\boldsymbol{\alpha}$, respectively, all feasible solutions of assignment matrix $\mathbf{Q}$ in \emph{transportation polytope}~\cite{cuturi2013sinkhorn} can be formulated as:
\begin{equation}
\scalebox{0.982}{$U(\boldsymbol{\alpha}, \boldsymbol{\beta}):=\left\{\mathbf{Q} \in \mathbb{R}^{K \times N} \mid \mathbf{Q} \mathbf{1}_N=\boldsymbol{\alpha}, \mathbf{Q}^{\top} \mathbf{1}_K=\boldsymbol{\beta}\right\}$},
\end{equation}
where $\mathbf{1}$ is an all-ones vector of corresponding dimension, and $\boldsymbol{\alpha}$ and $\boldsymbol{\beta}$ are the marginal projections of matrix $\mathbf{Q}$ onto its rows and columns, respectively.

Energy score from logit space is an effective and readily available metric to distinguish ID/OOD samples~\cite{energyood}, so we introduce the energy $\mathbf{e}\in\mathbb{R}^N$ as uncertainty guidance into our ET module. The energy of $x_i$ distributed across all clusters is shown as:
\begin{equation}\label{e:prior-energy} \\[-0.5ex]
e_i=\log\sum_{j=1}^K e^{l(c_j|x_i)}, 
\end{equation}
where $l(\mathbf{c}|x_i)=h_{ot}(z_i)$ denotes the logit of $x_i$ belonging to $c_j$. The energy $\mathbf{e}$ reflects the aggregate probability distribution of samples across clusters. Afterward, ET will encourage samples with higher energy scores to be assigned to the same cluster, while the ones with lower energy scores, which means they have larger uncertainty in the cluster distribution, will tend to be split uniformly among the $K$ clusters. And the matrix $\mathbf{Q}$ based on $\mathbf{e}$ can be denoted as:
\begin{equation}\label{e:r-c}
\boldsymbol{\alpha} = \frac{1}{K}\cdot\mathbf{1}_K,
\quad
\boldsymbol{\beta} = \frac{\mathbf{e}}{\sum_{i=1}^N e_i}.
\end{equation}
And the energy transport cost can be written as:  
\begin{equation}\label{e:energy_cost} 
\mathbf{P_{en}} = \mathbf{P}\cdot \mathbf{e} \in \mathbb{R} ^ {K \times N},
\end{equation}
where  $\mathbf{e}$ is first broadcast into a $K\times N$ matrix and then multiplied element-wise by $\mathbf{P}$. Hence, the Wasserstein distance between $\boldsymbol{\alpha}$ and $\boldsymbol{\beta}$ is defined as $\text{OT}(\boldsymbol{\alpha}, \boldsymbol{\beta})=\min _{\mathbf{Q} \in \Pi(\boldsymbol{\alpha}, \boldsymbol{\beta})}-{\langle \mathbf{Q}, \mathbf{P_{en}} \rangle}$, where $\langle\cdot, \cdot\rangle$ denotes the Frobenius dot-product. In this case, it is worth noting that $\mathbf{P_{en}}$ cannot be naively regarded as the conventional cost matrix in the optimal transport problem, since the elements in $\mathbf{P_{en}}$ represent the probability of a sample being assigned to a certain cluster. The larger the element value, the lower the cost of transferring the sample to the corresponding cluster.
To avoid solving this linear programming problem which needs a large computational cost, we introduce the entropic regularization term $\textit{H}(\mathbf{Q})$ into Wasserstein distance~\cite{cuturi2013sinkhorn} and express the optimization problem as:
\begin{equation}\label{e:entropic regularization}
   \text{OT}(\boldsymbol{\alpha}, \boldsymbol{\beta})=\min _{\mathbf{Q} \in \Pi(\boldsymbol{\alpha}, \boldsymbol{\beta})} -{\langle\ \mathbf{Q}, \mathbf{P_{en}}\rangle}+\varepsilon\textit{H}(\mathbf{Q}),
\end{equation}
where $\varepsilon >0$ and $\textit{H}(\mathbf{Q})=\sum_{ji}Q_{ji}\log Q_{ji}$. In this way, the optimal $\mathbf{Q}$ has been shown to be written as:
\begin{equation}\label{e:solution} 
\mathbf{Q}^{\frac{1}{\varepsilon}} =\text{Diag}(\mathbf{u})
        \mathbf{P_{en}}
        \text{Diag}(\mathbf{v}),
\end{equation}        
where exponentiation is element-wise, and $\mathbf{u}$ and $\mathbf{v}$ can be solved much faster by Sinkhorn’s Algorithm~\cite{cuturi2013sinkhorn}. The assignment matrix $\mathbf{Q}$ maps $N$ samples onto $K$ clusters.

Since then, we have obtained the cluster indexes of
all samples $\mathcal{C} = \{c_1,c_2,\dots,c_N\}\in \{1,2,\dots,K\}$. We collect unlabeled ID samples and assign pseudo labels to them according to the proportion of each label in a cluster \cite{yang2021semantically}. Specifically, we let samples that belong to $k$-th cluster at the $t$-th epoch form the set $D_k$ as denoted in ~\cref{e:Dk}.
\begin{equation}\label{e:Dk} 
    D_k^{(t)}=\{x_i\vert c^{(t)}_i=k\}.
	\end{equation}

 At epoch $t$, we define the set of ID-class labels $\mathcal{Y}^{(t)} = \mathcal{Y}_L \cup \mathcal{Y}_{ot}^{(t)} \in \{0,1,\dots,M-1\}$, where $M$ is the number of ID classes, and $\mathcal{Y}_L$ is formed by the ground-truth labels from $D_L$, while the ID pseudo labels assigned to unlabeled samples by ET constitute $\mathcal{Y}_{ot}^{(t)}$. Note that $\mathcal{Y}_{ot}^{(t)} = \varnothing$ when $t=0$ and it will be updated during training, while $\mathcal{Y}_L$ is unchanged.  Then we calculate the proportion of samples belonging to class $y \in \mathcal{Y}^{(t)}$ in cluster $k$ as follows:
\begin{equation}\label{E:class_rate} 
rate_{k,y}^{(t)}=\frac{\lvert D_{k,y}^{(t)}=\{x_i\vert c^{(t)}_i=k, y_i=y\}\rvert}{\lvert D_{k}^{(t)}\rvert}.
\end{equation}
When $rate_{k,c}^{(t)}$ is over a threshold $\tau$, all unlabeled samples in cluster $k$ will be added into labeled set $D_L$ to form the updated $D_{L}^{(t)}$ denoted as \cref{E:new_ind}, and $\mathcal{Y}^{(t)}$ is composed of the labels from $D_{L}^{(t)}$. 
\begin{equation}\label{E:new_ind}
	D_{L}^{(t)}= D_{L} \cup \{x_i \vert x_i \in D_k^{(t)}, rate_{k,y}^{(t)} > \tau\}.
\end{equation}
Meanwhile, the remaining unlabeled samples form $D_U$ at epoch $t$ is denoted as $D_{U}^{(t)}$. With the updated training set, the classification loss $L_{cls}^{(t)}$ and equalization loss  $ L_{unif}^{(t)}$ can be written as:
\begin{equation}\begin{split}
L_{cls}^{(t)} = -\sum_{x_i\in \mathcal{D}^{(t)}_L }\sum_{y_i\in \mathcal{Y}^{(t)}} 
 y_i \cdot \log (p(\mathbf{y}|x_i)), \\
L_{unif}^{(t)} = -\sum_{x_i\in\mathcal{D}^{(t)}_U} \frac{\mathbf{1}_M}{M} \cdot \log (p(\mathbf{y}|x_i)),
\end{split}\end{equation}
where $p(\mathbf{y}|x_i)$ means the prediction probability of $x_i$ over all ID classes, and $y_i$ should be converted to a one-hot vector denoting the corresponding label of $x_i$. Moreover, $\frac{\mathbf{1}_M}{M}$ denotes the uniform posterior distribution over all of $M$ ID classes and $L_{unif}^{(t)}$ will force samples in $D_{U}^{(t)}$ to uniformly distribute among $M$ ID classes~\cite{hendrycks18oe}.

\subsection{Inter-Cluster Extension Strategy}
Due to the fact that supervised learning only produces semantic representations that meet the minimum necessary for classification~\cite{con_ood}, to widen the discrepancy between ID/OOD on energy metric and thus strengthen the ability of energy transport cost in \cref{e:energy_cost} to guide ID/OOD samples to different clusters, we desire to obtain an improved feature representation by conducting an unsupervised training module called inter-cluster extension strategy.
 
With labels unavailable, given a batch of data $\{x_{(l,i)}\}_{i=1\dots B_1}$ and $\{x_{(u,i)}\}_{i=1\dots B_2}$ in $D_L$ and $D_U$ respectively, we combine them to form a batch of joint data $\{x_i\}_{i=1\dots B}=\{x_{(l,i)}\} \cup \{x_{(u,i)}\}$, where  $B=B_1+B_2$, Then we obtain two augmentation versions of them denoted as $x_i^0=A^0(x_i)$ and $x_i^1=A^1(x_i)$ through two kinds of data augmentations $A^0$ and $A^1$, respectively. Then the augmented data will be fed into an encoder to be extracted feature as:
\begin{equation}\label{e:rep}
\begin{split}
    z_i^0 = h_m(e_\theta^0(x_i^0)), \\
    z_i^1 = h_m(e_\theta^1(x_i^1)),
\end{split}    
\end{equation} 
where $e_\theta^0$ and $e_\theta^1$ are two encoders with parameters $\theta$, and $h_m$ is an MLP. Additionally, a memory queue is created to store $z_i^1$ at different iterations and the $n$ latest batches of $z_i^1$ compose a dynamic queue. The dynamic queue at iteration $t'$ is described as: 
\begin{equation}\label{e:queue}
    Z^{1,(t')} = \{z_i^1\}^{(t')} \cup \{z_i^1\}^{(t'-1)}  {\dots}  \cup\{z_i^1\}^{(t'-n+1)}.
\end{equation}
Considering the absence of labels, we obtain the desired representation by maximizing the cosine similarity of the same samples in $\{z_i^0\}^{(t')}$ and $\{z_i^1\}^{(t')}$, so we employ InfoNCE loss as the objective:
\begin{equation}
L_{rep}=-\sum_{i=1}^N \log \frac{\exp(cos(z_i^0, z_i^1))}{\sum_{j=1}^{n \cdot B}\exp(cos(z_i^0, z_j))},
\end{equation}
where $z_j$$\in$$Z^{1,(t')}$ and $cos$ denotes cosine similarity. The enhanced representation obtained by $L_{rep}$ will be mapped to the lower-dimensional logit space, where we can obtain the class-specific energy score better reflecting the semantic discrepancy of samples from different classes and assign a more discriminative cluster distribution to them.

Our overall objective can be expressed as~\cref{e:overall_loss} with the wights $\gamma$ and $\lambda$.
\begin{equation}\label{e:overall_loss}
L = L_{cls}^{(t)} + \gamma L_{unif}^{(t)} + \lambda L_{rep}.
\end{equation}

\subsection{T-Energy Score}\label{subsec: OOD metric}
Considering the parallel structure of $h_{ot}$ and $h_{cls}$, which both map the same features to the logit space through different fully-connected layers, there is a certain degree of correlation between the two logits. So we use the energy score from the logit $l(\mathbf{y}|x_i)=h_{cls}(z_i)$ outputted by $h_{cls}$ as OOD score to detect OOD samples after training. The energy can be readily calculated through the \verb'logsumexp'\ operator as $energy(x_i)=\log\sum_{m=1}^M e^{l(y_m|x_i)}$, where $l(y_m|x_i)$ is the logit of $x_i$ belonging to the ID class $y_m$, and ID samples have higher energy score. We observe that OOD samples are more concentrated at minimum energy compared to methods training with only ID samples (\textit{e.g.}, \cite{energyood}) due to the involvement of training OOD data in SCOOD. In this case, a large temperature value $T$ can smooth the energy distribution and thereby widen the discrepancy between ID/OOD on the energy metric.
Therefore, we implement the temperature scaling proposed in ODIN~\cite{odin} to compute energy score using temperature-scaled logits instead of mapping them to softmax score like ODIN, which is consistent with the source of the uncertainty metric in the proposed ET. The T-energy score is formulated as:
\begin{equation}\label{e:T-energy}
T \mbox{-} energy(x_i)
=
T \cdot \log\sum_{m=1}^M e^{l(y_m|x_i) / T}.
\end{equation}

\section{Experiments}\label{sec:exp}

\begin{table*}[btp!]
\renewcommand{\arraystretch}{1.}
\renewcommand{\tabcolsep}{8.pt}
\centering
\small
\caption{\textbf{Comparison between the previous SOTA methods and ours on the two SCOOD benchmarks.} We report the averaged values on 6 OOD datasets and detailed results are shown in Appendix. $\uparrow$/$\downarrow$ indicates higher/lower value is better. The best results are in \textbf{bold}.}
\label{T:Results}
\begin{tabular}{cc|ccc|cccc}
\toprule
\multirow{2}{*}{Benchmarks} 
& \multirow{2}{*}{Method} 
& \multirow{2}{*}{FPR95~$\downarrow$} 
& \multirow{2}{*}{AUROC~$\uparrow$} 
& \multirow{2}{*}{AUPR-In/Out~$\uparrow$}
& \multicolumn{4}{c}{CCR@FPR~$\uparrow$} \\ 
\cmidrule(lr){6-9}
& && && $10^{-4}$  & $10^{-3}$  & $10^{-2}$ & $10^{-1}$                      \\ \midrule
\multirow{6}{*}{\begin{tabular}[c]{@{}c@{}}CIFAR-10\\ Benchmark\end{tabular}}
& ODIN~\cite{odin}         &  52.00 & 82.00 & 73.13~/~85.12 & 0.36 & 1.29& 	6.92    & 39.37 \\ 
& EBO~\cite{energyood}      &  50.03 &	83.83 &	77.15~/~85.11 & 0.49	& 1.93	&9.12	&46.48 \\
& OE~\cite{hendrycks18oe}  & 50.53	& 88.93	& 87.55~/~87.83	& 13.41	& 20.25	& 33.91	& 68.20 \\
& MCD~\cite{mcd}           &73.02&	83.89&	83.39~/~80.53&	5.41&	12.3&	28.02&	62.02  \\
& UDG\cite{yang2021semantically}   & 36.22 & 93.78 & 93.61~/~92.61 & 13.87  & 34.48 & 59.97 & 82.14 \\
& \textbf{Ours}     & \textbf{8.53} & \textbf{96.47} & \textbf{97.10}~/~\textbf{95.65}  &  \textbf{40.31} & \textbf{63.95} & \textbf{77.35} & \textbf{86.27}          \\ 
\midrule
\multirow{6}{*}{\begin{tabular}[c]{@{}c@{}}CIFAR-100\\ Benchmark\end{tabular}}
& ODIN~\cite{odin}        &81.89	&77.98	&78.54~/~72.56	&1.84	&5.65	& 17.77	& 46.73 \\
& EBO~\cite{energyood}    &81.66	&79.31	&80.54~/~72.82	&2.43	&7.26	& 21.41	& 49.39 \\
& OE~\cite{hendrycks18oe} &80.06	&78.46	&80.22~/~71.83	&2.74	&8.37	& 22.18	& 46.75	\\
& MCD~\cite{mcd}          & 85.14	&74.82	&75.93~/~69.14  &1.06   & 4.60  & 16.73 & 41.83 \\
& UDG~\cite{yang2021semantically}   & 75.45 & 79.63  & 80.69~/~74.10  &  3.85  & 8.66 &  20.57 &  44.47           \\ 
& \textbf{Ours}     & \textbf{41.05} & \textbf{82.44}  & \textbf{84.37}~/~\textbf{76.47}  &  \textbf{11.70}  & \textbf{19.47} &  \textbf{34.05} &  \textbf{50.97}   \\ 
\bottomrule
\end{tabular}
\end{table*}

\subsection{Benchmarks} We evaluate our method on the realistic SCOOD benchmarks~\cite{yang2021semantically} proposed recently. SCOOD benchmarks contain two benchmarks: CIFAR-10 benchmark and CIFAR-100 benchmark which regard CIFAR-10 and CIFAR-100~\cite{cifar} as the labeled set $D_L$, respectively. Tiny-ImageNet~\cite{tinyImageNet} mixed with ID and OOD data is utilized as extra unlabeled training set $D_U$ on SCOOD benchmarks. In test time, CIFAR-100 is used as one of the OOD datasets for CIFAR-10 benchmark while CIFAR-10 is included in the OOD set of CIFAR-100 benchmark. Additionally, five other datasets including Texture~\cite{texture}, SVHN~\cite{svhn}, Tiny-ImageNet~\cite{tinyImageNet}, LSUN~\cite{lsun} and Places365~\cite{places365} are collected to form the complete test set $T$ of the two benchmarks, which contain both ID and OOD samples. SCOOD benchmarks re-split these samples into $T^I$ and $T^O$ according to their true semantics. 

\subsection{Evaluation Metrics.} Following UDG~\cite{yang2021semantically}, we use five metrics to evaluate the performance of our method. 
\textbf{FPR@TPR95$\%$} presents the ratio of falsely identified OOD when most (\textit{i.e.}95$\%$) ID samples are correctly recognized.
\textbf{AUROC} is the area under the receiver operating characteristic curve, evaluating the OOD detection performance.
\textbf{AUPR-In/Out} calculates the area under the precision-recall curve. AUPR-In/OUT denotes ID/OOD samples as positive.
\textbf{CCR@FPR$n$} shows classification accuracy when the ratio of falsely identified OOD equals $n$, which can evaluate ID classification and OOD detection capability simultaneously.

\begin{table}[btp!]
\renewcommand{\arraystretch}{1.}
\renewcommand{\tabcolsep}{1.pt}
\small
\caption{\textbf{Ablation study on CIFAR-10 benchmark.} We take UDG~\cite{yang2021semantically} as baseline method. Our completed method combining ET and $L_{rep}$ is written as `Ours' for short. All experiments are tested in the T-energy score mentioned in~\cref{subsec: OOD metric} for fair comparisons. For brevity, we refer to each experiment by its index.}
\label{T:Ablation}
\centering
\begin{tabular}{c|c|cccc}
\toprule
& Strategy & FPR95~$\downarrow$ & AUROC~$\uparrow$ & AUPR-In/Out~$\uparrow$ & ACC~$\uparrow$ \\  \midrule
1: & UDG \cite{yang2021semantically}  & 21.57 & 92.44 & 91.70~/~92.12  & 92.28  \\
2: & UDG + $L_{rep}$  & 17.75 & 94.77 & 94.80~/~93.13 & 93.07\\
3: & ET     &  11.39 & 96.05 & 96.18~/~94.57 & 93.40    \\ 
4: & \textbf{Ours}  & \textbf{8.53} & \textbf{96.47} & \textbf{97.10}~/~\textbf{95.65} &  \textbf{93.71}           \\ 
\bottomrule
\end{tabular}
\end{table}

\begin{figure}[t]
\centering
\begin{overpic}[width=0.99\linewidth]{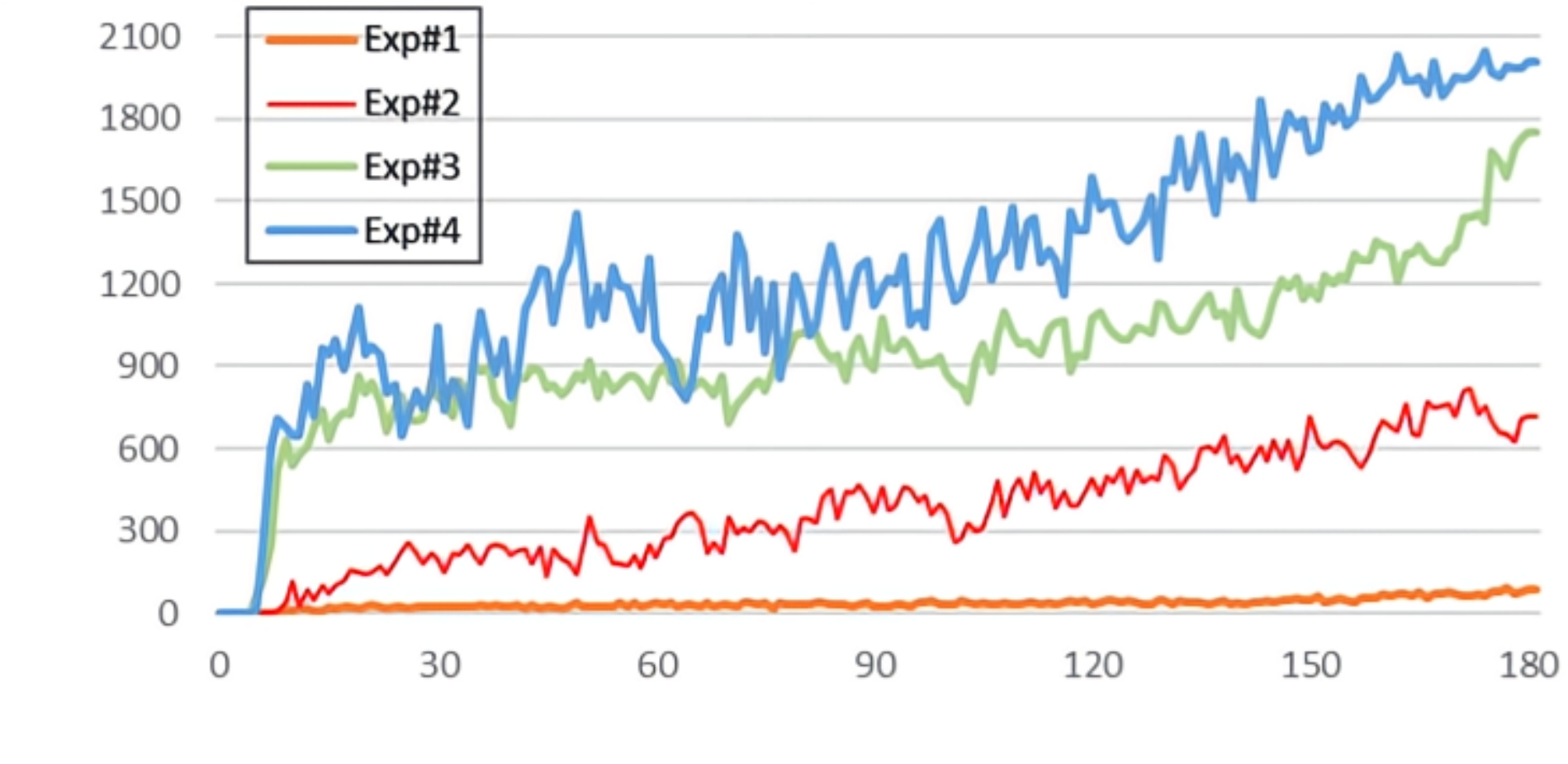}
 \put(50,2){\textbf{\footnotesize{Epochs}}}
  \put(2, 26){\rotatebox{90}{\footnotesize\textbf{Num}}}
\end{overpic}
\vspace{-4pt}
\caption{\textbf{The number of accurately assigned labels during training in EXP$\#1$ \mbox{--} $\#4$} shows that the ET has a significant improvement over IDF strategy in assigning exact labels to unlabeled ID samples, and $L_{rep}$ can further increase this advantage. EXP$\#1$ is the UDG baseline, EXP$\#2$ introduces the $L_{rep}$ into UDG, EXP$\#3$ is the proposed ET alone, and EXP$\#4$ is the complete version of our method.}
\label{Fig:num}
\end{figure}

\subsection{Results on SCOOD Benchmarks} We compare the results of our proposed approach with the previous state-of-the-art (SOTA) OOD detection methods in \cref{T:Results}. OIDN~\cite{odin} and EBO~\cite{energyood} do not require extra unlabeled training data, and OE~\cite{hendrycks18oe}, MCD~\cite{mcd}, and UDG~\cite{yang2021semantically} use Tiny-ImageNet as unlabeled training data $D_U$. All experiments use ResNet-18~\cite{resnet}. We only report the average metric values on 6 OOD test datasets for each benchmark limited by space. Results show that our proposed uncertainty-aware optimal transport scheme consistently obtains the best results across all metrics, especially FPR@95 is significantly improved on both benchmarks. In particular, although MCD~\cite{mcd} introduces extra unlabeled training data and achieves satisfying results across traditional OOD benchmarks, its maximization of the entropy discrepancy between ID/OOD samples makes the model pay excessive attention to the low-level covariate shifts, which eventually leads to limited generalization in other OOD sources and failure on the SCOOD task.

\subsection{Ablation Study and Qualitative Analysis}

\textbf{Effectiveness of uncertainty-aware optimal transport scheme.}  Here we analyze the effect of each major component, including the energy-based transport (ET) mechanism and inter-cluster extension strategy $L_{rep}$ in \cref{T:Ablation}. EXP$\#1$ is the UDG~\cite{yang2021semantically} method which we use as the baseline, and then we introduce the $L_{rep}$ into UDG to carry out EXP$\#2$. To evaluate the effectiveness of the proposed ET, we replace the IDF strategy in UDG with ET and implement EXP$\#3$, and EXP$\#4$ which combines ET with the $L_{rep}$ is the complete version of our method.

From the comparison between EXP$\#2$ and EXP$\#1$, it can be seen when $L_{rep}$ is introduced in UDG and combined with IDF strategy~\cite{yang2021semantically} based on K-means clustering, the result is improved compared with the baseline. It is intuitive that an enhanced feature representation produced by $L_{rep}$ increases the ID/OOD discrepancy and facilitates clustering-based label assignment in the feature space. However, limited by the non-robust feature distance to covariate shifts and lacking the practical guidance in clustering task, simply applying $L_{rep}$ in SCOOD fails to overcome the impact of covariate shift, so the IDF strategy still has shortcomings in extracting ID semantics from the unlabeled set, as proved in \cref{Fig:num}. When we replace the IDF strategy with the ET based on the energy uncertainty and implement the Exp$\#3$, the result shows that the OOD detection ability is enhanced by a large margin, as FPR@95 gets a significant 10.18$\%$ improvement than baseline. After combining ET and $L_{rep}$, the results of EXP$\#4$ show that the performance of the model is at its best when obtaining a more discriminative uncertainty derived from a enhanced representation to promote the discrimination of inter-cluster distribution. In \cref{Fig:num} we also report the comparison among EXP$\#1$ \mbox{--} $\#4$ in the number of accurately assigned labels, we can see from where both the ET and $L_{rep}$ have brought apparent ascension in the aspect of increasing the accuracy of the label assignment. Combined with the results in \cref{T:Ablation} and \cref{Fig:num}, it is again demonstrated that more accurate semantic label assignment during training will promote the performance of the model, since the model benefits from exploring the semantic discrepancy and knowledge hidden in the unlabeled set more sufficiently.


In summary, the performance of the model boosts mainly due to the contribution of ET, which provides effective guidance for the cluster distribution of samples based on semantics via energy metric. Then $L_{rep}$ further enlarges the advantages over baseline by providing a more discriminate energy metric between ID/OOD.

\begin{figure}[t]
\centering
\begin{overpic}[width=0.99\linewidth]{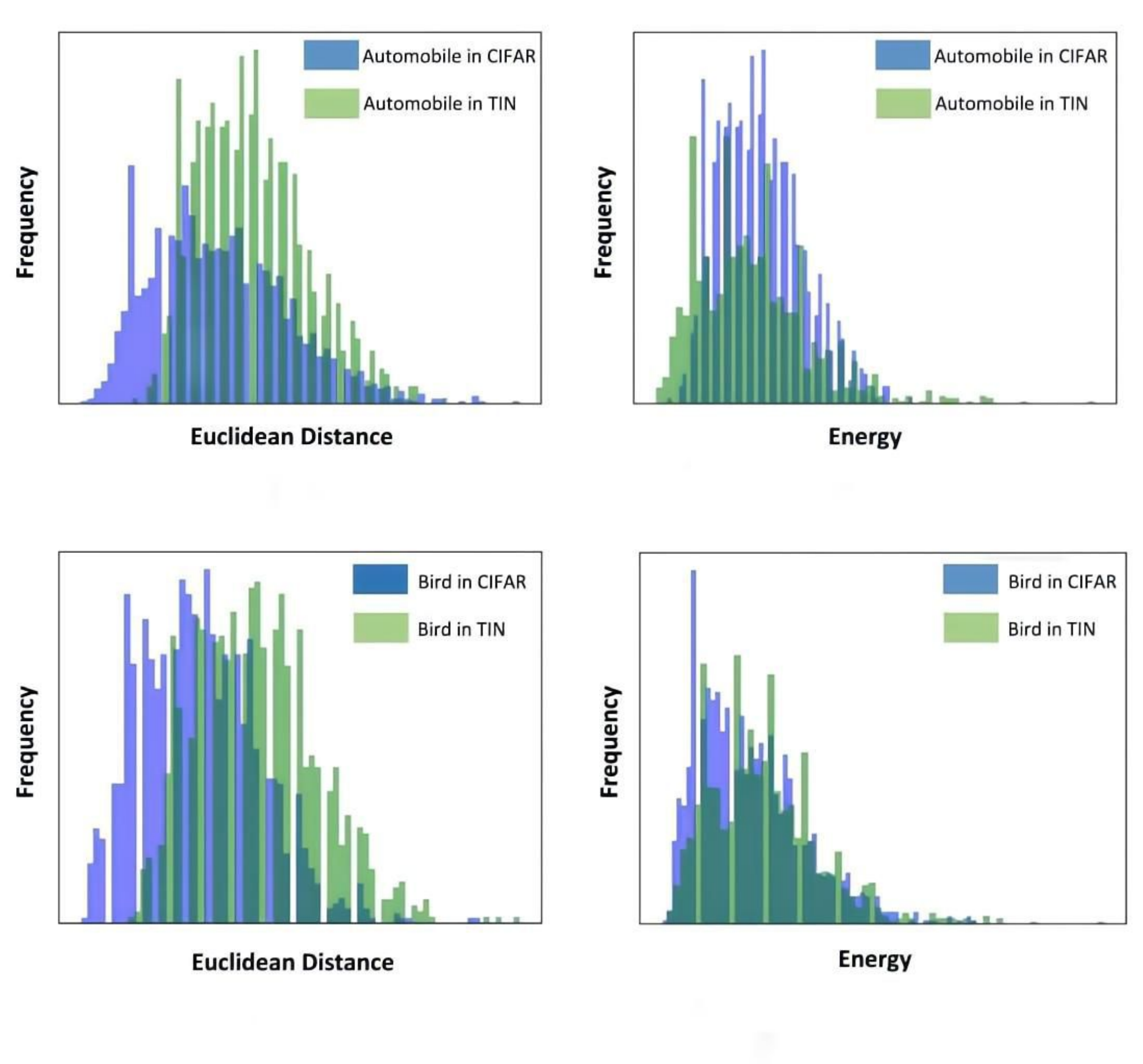}
\put(10.5, 50){\footnotesize{(a) Distance vs Energy distributions of `automobile' images}}
  \put(15, 4){\footnotesize{(b) Distance vs Energy distributions of `bird' images}}
\end{overpic}
\vspace{-5pt}
\caption{\textbf{The effectiveness of energy metric.} In `automobile' and `bird' classes, the energy metric still maintains better consistency with the interference of covariate shifts. CIFAR and TIN denote CIFAR-10 and Tiny-ImageNet datasets, respectively.}
\label{Fig:consistency}
\end{figure}

\begin{table}[t]
\centering
\renewcommand{\arraystretch}{1.}
\renewcommand{\tabcolsep}{4.pt}
\small
    \caption{\textbf{The effectiveness of energy in ET.} V-OT and En-OT denotes the methods introducing \emph{variance} and \emph{entropy} as uncertainty like ET, respectively. The performance of our proposed ET outperforms the two methods on all metrics.}
    \label{T:OT}
    \begin{tabular}{c|cccc}
    	\toprule
    	 & FPR95~$\downarrow$ & \footnotesize{AUROC~$\uparrow$} & \footnotesize{AUPR-IN/OUT~$\uparrow$} & \footnotesize{ACC~$\uparrow$} \\
    	\midrule
    	V-OT                   & 13.46 & 96.11 & 96.70~/~95.43 & 93.15 \\
    	En-OT                 & 10.94 & 95.86 & 96.75~/~94.67 & 93.10 \\
    	\textbf{ET}             & \textbf{8.53} & \textbf{96.47} & \textbf{97.10}~/~\textbf{95.65} &  \textbf{93.71}   \\
    	\bottomrule
    \end{tabular}
\end{table} 

\textbf{The consistency and effectiveness of energy metric in ET.} 
In this work, we propose the energy-based transport (ET) mechanism to explore the semantic discrepancy and knowledge hidden in the unlabeled set more sufficiently. 
Here, we first demonstrate the robustness of the energy metric to covariate shifts on more classes. We employ a model trained only with CIFAR-10 to output the minimum Euclidean distance and energy distributions for samples belonging to the same class but from different datasets (\textit{e.g., }CIFAR-10 and Tiny-ImageNet). Except for the distributions of the `dog' images shown in the histograms of \cref{fig:motivation}, we perform the same experiment with the `automobile' and `bird' images and report the distribution difference in \cref{Fig:consistency}. In `automobile' and `bird' categories, the energy metric still better overcomes the interference of the covariate shifts and maintains the consistence of the semantic compared with feature distance. Afterward, we consider other commonly-used uncertainty metrics to replace the energy metric, including the \emph{variance} and \emph{entropy} of samples over cluster distribution, and replace the energy metric in ET with them to guide label assignment. The result in \cref{T:OT} indicates that our proposed ET outperforms both methods based on other uncertainty, confirming the effective guidance of energy in the assignment process.


\textbf{Effectiveness of extra data and classification accuracy decay.}
The comprehensive survey~\cite{yang2021generalized} indicates that extra data generally promotes the OOD detection performance of models, but the classification performance of classification-based OOD detection methods will inevitably decrease due to the interferences of OOD samples. To explore the impact of the extra training data on the performance of OOD detection and ID classification, we first evaluate the model only trained on CIFAR-10 with standard cross-entropy loss (CE loss) and then introduce Tiny-ImageNet as extra unlabeled training data and implement EXP$\#2$ \mbox{--} $\#4$. The results are reported in \cref{T:Extra Data}.

\begin{table}[t]
\centering
\renewcommand{\arraystretch}{1.}
\renewcommand{\tabcolsep}{1.pt}
    \small
    \caption{\textbf{The impact of extra training data.} EXP$\#1$ uses a model trained only on CIFAR-10 (CIFAR) with CE loss, while EXP$\#2$ \mbox{--} $\#4$  introduce Tiny-ImageNet (TIN) as extra unlabeled data and distinguish unlabeled ID/OOD samples in training. Extra training data consistently promotes the OOD detection performance but also brings a drop to classification accuracy (ACC). Our method (including both ET and $L_{rep}$, written as `Ours' for short) minimizes the degradation on ACC.}
    \label{T:Extra Data}
    \centering
    \begin{tabular}{c@{\hskip 8pt}l|@{\hskip 1pt}c@{\hskip 1pt}c@{\hskip 1pt}c@{\hskip 1pt}@{\hskip 1pt}c}
    	\toprule
    	$D$ & Strategy & \footnotesize{FPR95~$\downarrow$} & \footnotesize{AUROC~$\uparrow$} & \footnotesize{AUPR-IN/OUT~$\uparrow$} & \footnotesize{ACC~$\uparrow$} \\
    	\midrule
    	{\footnotesize{CIFAR}}     
    	& 1: CE loss  & 34.96 & 85.72 & 84.58~/~83.24 & 94.94   \\
    	\midrule
    	\multirow{3}{*}{\begin{tabular}[c]{@{}c@{}}\footnotesize{CIFAR}\\\footnotesize{+}\\\footnotesize{TIN}\end{tabular}}
    	& 2: UDG\cite{yang2021semantically}   & 21.57 & 92.44 & 91.70~/~92.12  & 92.28  \\
    	&  3: ET                     &  10.21 & 96.25 & 96.92~/~94.76 & 93.48  \\
    	& 4: \textbf{Ours}    & \textbf{8.53} & \textbf{96.47} & \textbf{97.10}~/~\textbf{95.65} &  \textbf{93.71}  \\
    	\bottomrule
    \end{tabular}
\end{table}

By comparing EXP$\#2$ \mbox{--} $\#4$ and EXP$\#1$, we can see when the model is exposed to the extra unlabeled set mixed with ID and OOD and distinguishes the unlabeled ID/OOD samples during training as UDG and the ET do, the OOD detection performance will be comprehensively improved because the model keeps focusing on the semantic shifts between ID/OOD thus learning their discrepancy better. However, since unlabeled ID samples cannot be collected as fully as ideal, some of them are classified into the wrong categories, which hurts the classification performance of the model. Our method brings a more accurate label assignment derived from the practical guidance of energy score on the cluster distribution, and mitigates accuracy decay induced by extra OOD samples compared to UDG.

\textbf{Influence of OOD scores.} We notice that when replacing max softmax probability (MSP)~\cite{baseline} with T-energy as OOD score, UDG drops by about 15$\%$ on FPR@95. To exclude the possibility that our approach only gains from T-energy score, we test the performance of UDG and our approach on three kinds of OOD scores, including MSP, energy and T-energy. As shown in \cref{T:metrics}, our method outperforms the baseline no matter which OOD score is chosen. In particular, when evaluating UDG on energy and T-energy, the AUROC and AUPR both suffer a drop. However, the proposed ET explores the ID/OOD semantic discrepancy based on the energy, and in turn promotes the ability of energy score to discriminate ID/OOD by continuously optimizing label assignment, so our method obtains improved performance when being tested with energy score. Additionally, temperature scaling further widens the gap between ID and OOD, helping the model achieves the best performance on all metrics.

\begin{table}[t]
\centering
\renewcommand{\arraystretch}{1.}
\renewcommand{\tabcolsep}{3.pt}
\small
    \caption{\textbf{Influence of OOD scores.} MSP denotes max softmax probability. Our method outperforms the UDG on all OOD scores and achieves the best on the T-energy.}
    \label{T:metrics}
    \centering
    \begin{tabular}{cc|ccc}
    	\toprule
    	OOD Score & Method & \footnotesize{FPR95~$\downarrow$} & \footnotesize{AUROC~$\uparrow$} & \footnotesize{AUPR-IN/OUT~$\uparrow$} \\
    	\midrule
    	\multirow{2}{*}{MSP}        
    	& UDG\cite{yang2021semantically}                   & 36.22 & 93.78 & 92.61~/~92.94 \\
    	& Ours                 & 13.55 & 95.56 & 96.89~/~94.96\\
    	\midrule
    	\multirow{2}{*}{Energy}        
    	& UDG\cite{yang2021semantically}                   & 34.90 & 90.65 & 91.56~/~91.11 \\
    	& Ours                 & 12.86 & 96.05 & 97.04~/~95.01\\
    	\midrule
    	\multirow{2}{*}{T-Energy}        
    	& UDG\cite{yang2021semantically}                   & 21.57 & 92.44 & 91.70~/~92.12 \\
    	& Ours                 & \textbf{8.53} & \textbf{96.47} & \textbf{97.10}~/~\textbf{95.65}\\
    	\bottomrule
    \end{tabular}
\end{table}

\section{Conclusion}\label{sec:conclusion}
In this work, we focus on addressing the fundamental challenge of OOD detection tasks and further explore how to fully understand the semantic discrepancy between the ID/OOD samples on SCOOD benchmarks. We reveal that the key to success in the realistic SCOOD task is to allocate as many ID samples in the unlabeled set correctly as possible. To this end, we propose a novel uncertainty-aware optimal transport scheme, which introduces class-specific energy scores as guidance for effective label assignment. Experimental results show that our method achieves better performance than previous state-of-the-art methods on SCOOD benchmarks.

\textbf{Limitations.} In addition to temperature scaling, other techniques such as feature clipping applied in ReAct~\cite{sun2021react} also enhance the performance of energy score, so how to obtain an OOD score that best fits the SCOOD task can be further explored. Moreover, a setting highly related to SCOOD has been proposed in \cite{katz2022training} and formulated as a constrained optimization problem. We will also theoretically analyze these practical OOD settings in our feature work.

\textbf{Acknowledgments.} 
This work is supported by National Key R\&D Program of China under Grant 2020AAA0105701 and Ant Group through Ant Research Intern Program.

\newpage
{\small
\bibliographystyle{ieee_fullname}
\bibliography{ref}
}

\clearpage

\appendix

\section*{Appendix}

\section{Experiment Details}
The ResNet-18~\cite{resnet} is employed as the backbone for all experiments, which is trained by an SGD optimizer with a weight decay of $0.0005$ and a momentum of $0.9$. We use the cosine annealing learning rate starting at $0.1$, taking totally $180$ epochs. Two dataloaders are prepared with batch-size of $64$ and $128$ for $\mathcal{D}_L$ and $\mathcal{D}_U$, respectively.
For the objective of training is denoted as:
\begin{equation}\label{e:object}
L = L_{cls}^{(t)} + \gamma L_{unif}^{(t)} + \lambda L_{rep},
\end{equation}
where we set $\gamma=0.5$ and $\lambda=0.3$ for all experiments. The number of cluster $K$ for CIFAR-10/100 benchmark is $1024$/$2048$.

\section{Discussion of Training Process} \label{subsec:process}

In summary, we alternate the following two steps throughout the training process:

\noindent \textbf{1: Representation learning.} 
Given the updated $D_{L}^{(t)}$ and $D_{U}^{(t)}$ based on the assignment matrix $\mathbf{Q}$, the model is trained with \cref{e:object} including the inter-cluster extension strategy $L_{rep}$ to obtain a discriminative representation between each ID class and OOD class.

\noindent \textbf{2: Optimizing label assignment.} We fix the parameters of the model, and use the model to estimate the energy-based transport cost in the proposed energy-based transport (ET) mechanism. Then we employ the ET to assign correct labels to unlabeled ID samples as many as possible to optimize the assignment matrix $\mathbf{Q}$ with the guidance of the energy-based transport cost.

\begin{figure}[t]
\centering
\begin{overpic}[width=0.99\linewidth]{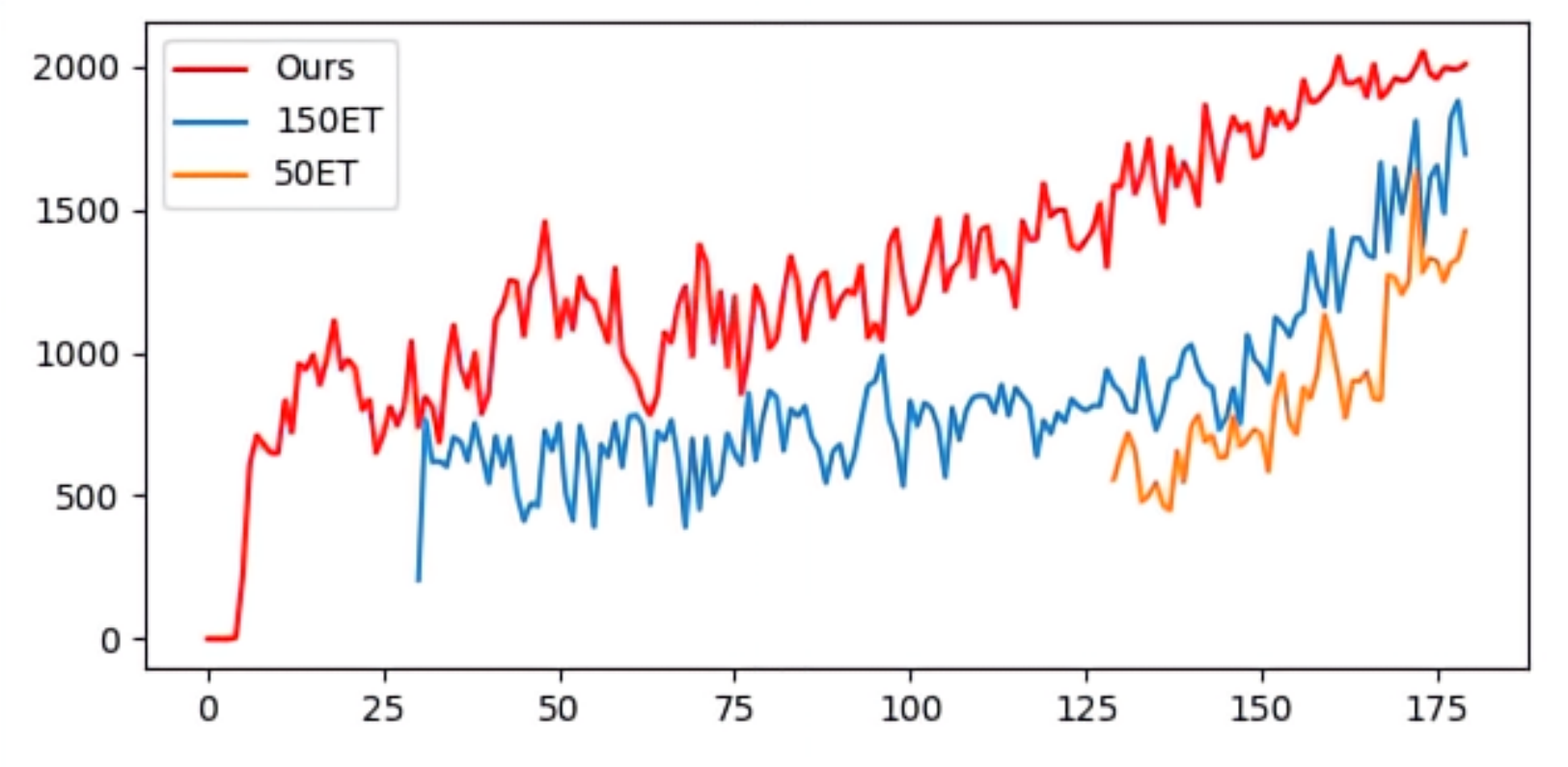}
 \put(48,0){\textbf{\footnotesize{Epochs}}}
  \put(-1,23){\rotatebox{90}{\footnotesize\textbf{Num}}}
\end{overpic}
\caption{\textbf{The number of accurately assigned labels during training in the three experiments in \cref{T:process}} shows that our method significantly improves over the other two strategies in assigning exact labels to unlabeled ID samples.`50ET' denotes the model performs  the two steps mentioned in \cref{subsec:process} alternately only for the last 50 epochs, while `150ET' alternates the two steps for the last 150 epochs. `Ours' means our method which alternates the two steps throughout the training process.}
\label{Fig:process}
\end{figure}

Notably, the ET is performed at the beginning of the training. Limited by the  representation not strong at this stage, the energy metric may not reflect the discrepancy in ID/OOD, thus providing ineffective guidance or even accumulating errors. To explore this doubt, we trained the model only using \cref{e:object} for the firstly and performed the above two steps alternately at the remaining epochs, and then evaluated these strategies in \cref{T:process}. Results show that our method obtains consistently best results across all metrics, which means that performing ET during the whole training process can more fully learn the discrepancy in ID/OOD. In \cref{Fig:process} we also report the comparison among the three experiments above in the number of accurately assigned labels. It can be seen that our method no matter at which epoch can allocate more accurate labels than `50ET' and `150ET', so the ID semantic knowledge hidden in unlabeled set mined by ET at the first few epochs is beneficial to the subsequent training, and the strategy which our method adopts ultimately converges and works.

\begin{figure}[t]
\centering
\begin{overpic}[width=0.99\linewidth]{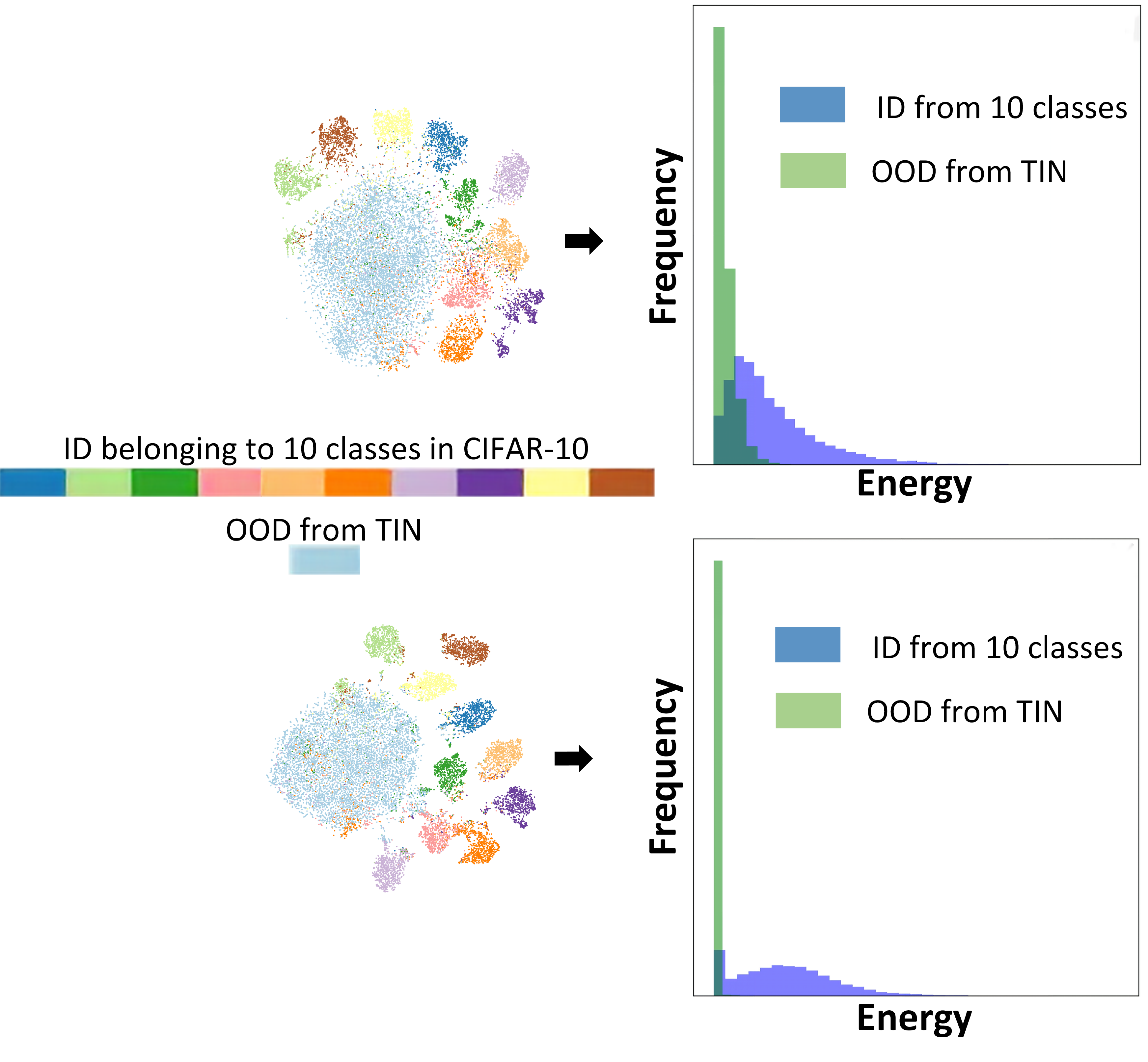}
 \put(0, 73){\scriptsize{Feature Outputted}}
  \put(0, 69){\scriptsize{by Last Layer }}
  \put(0, 65){\scriptsize{\textbf{Without} $L_{rep}$}:}
  \put(0, 30){\scriptsize{Feature Outputted}}
  \put(0, 26){\scriptsize{by Last Layer }}
  \put(0, 22){\scriptsize{\textbf{With} $L_{rep}$}:}
\end{overpic}
\caption{\textbf{Comparison of feature representations and the energy metric introduced in the ET between using $L_{rep}$ and without $L_{rep}$.} $L_{rep}$ not only produces more distinguishable and compact representations, but more importantly, energy metric that reflects ID/OOD differences more significantly can be obtained. From the statistical histogram, it can be seen that when $L_{rep}$ is used, the energy of OOD samples is concentrated at the minimum value and has little overlap with the distribution of ID. These OOD samples will be forced to be uniformly distributed over all clusters in ET. TIN denotes the Tiny-ImageNet dataset.}
\label{Fig:rep_energy}
\end{figure}

\begin{figure}[t]
\centering
\begin{overpic}[width=0.99\linewidth]{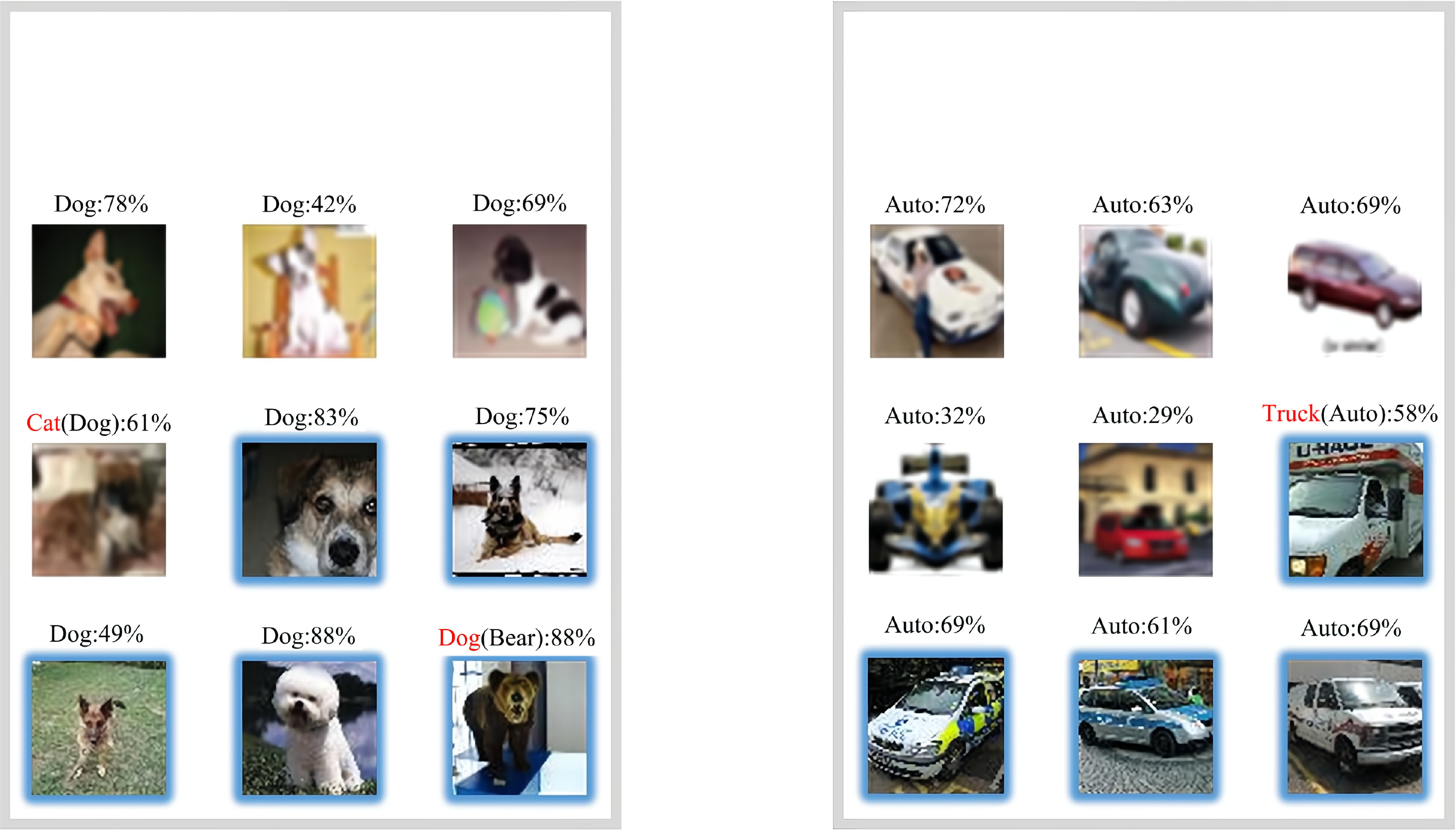}
 \put(7,50){\footnotesize{A Cluster Dominated}}
 \put(15,47){\footnotesize{by `Dog'}}
  \put(63,50){\footnotesize{A Cluster Dominated}}
  \put(62,47){\footnotesize{ by `Automobile(Auto)'}}
\end{overpic}
\caption{\textbf{Visualization of the proposed ET.} We show partial samples from two clusters where the proportion of "Dog" and "Automobile" classes exceed 75$\%$, respectively. The predicted labels and the corresponding prediction probability produced by the model are noted above each image. The \textcolor{red}{red} predicted label means that the model classifies the image into a wrong class, and its ground-truth label is in parentheses. The images with blue edges are from the unlabeled Tiny-ImageNet dataset, and the other are from CIFAR-10. The visualization shows that our ET is capable of assigning correct semantic labels to unlabeled ID samples incorrectly predicted by the model or with low confidence.}
\label{Fig:ET}
\end{figure}
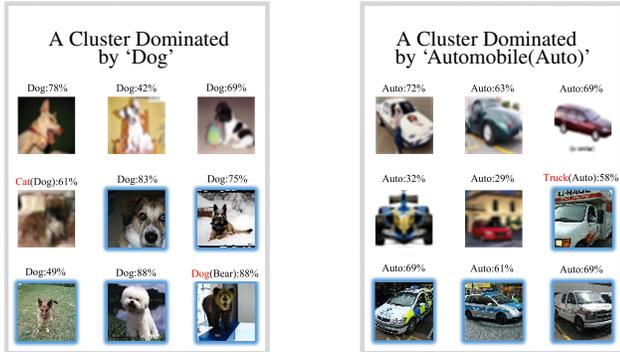

\begin{table}[btp!]
\centering
\renewcommand{\arraystretch}{1.}
\renewcommand{\tabcolsep}{6.pt}
    \small
\caption{\textbf{Comparison between different strategies of training process.} `50ET' denotes the model is trained only with \cref{e:object} for the first 130 epochs and performed the two steps mentioned in \cref{subsec:process} alternately for the last 50 epochs. While `150ET' uses \cref{e:object} for training for the first 30 epochs, and performs the two steps alternately for the last 150 epochs. `Ours' means our method which alternates the two steps throughout the training process. $\uparrow$/$\downarrow$ indicates higher/lower value is better. The best results are in \textbf{bold}.}
\label{T:process}
\begin{tabular}{c|ccc|c}
\toprule
{Strategy} 
& \footnotesize{FPR95~$\downarrow$} 
& \footnotesize{AUROC~$\uparrow$} 
&  \footnotesize{AUPR-In/Out~$\uparrow$}
&  \footnotesize{ACC~$\uparrow$} \\ 
\midrule
50ET & 13.54 & 93.54 & 94.72~/~93.53 &  91.97 \\
150ET & 11.83 & 96.19 & 95.86~/~94.22 & 92.87\\
\textbf{Ours}     & \textbf{8.53} & \textbf{96.47} & \textbf{97.10}~/~\textbf{95.65}  & \textbf{93.71}  \\ 
\bottomrule
\end{tabular}
\end{table}

\section{Effectiveness of the $L_{rep}$}

The inter-cluster extension strategy ($L_{rep}$) enhances the global feature representation mixed with ID and OOD samples and then the enhanced representation will be mapped into a more discriminate logit space. The energy metrics produced in this space can better reflect the ID/OOD differences to more effectively guide the cluster distribution of ID/OOD samples. In \cref{Fig:rep_energy}, we use TSNE~\cite{van2008visualizing} to visualize the learned feature representation and compare the energy metric of ID/OOD in training set with or without $L_{rep}$. This figure demonstrates the contribution of $L_{rep}$ to the ability of energy metric in the ET to reflect the discrepancy between ID/OOD. Benefiting from the effective guidance of this energy metric for samples with different semantics, $L_{rep}$ ultimately further facilitates ET to explore semantic knowledge hidden in the unlabeled set and improve the performance of the model.

\section{Visualization of the ET}

Considering the overconfident prediction of deep neural network models on OOD inputs revealed in \cite{nguyen2015deep,scheirer2012toward}, we cannot assign labels to unlabeled samples relying on the predicted results of the network. Moreover, in the experiment we found that the network will predict the ID samples into wrong classes or output insignificant confidence (softmax probability) on the correct classes. We demonstrate the superiority of the proposed ET in assigning accurate labels through the visualization in \cref{Fig:ET}, it can be seen from where that our ET can collect unlabeled ID images in a correct manner. This strategy splits the ID samples incorrectly predicted by the model (refer to the two images being predicted into `\textcolor{red}{cat}' and `\textcolor{red}{truck}') or with low confidence (around 30$\%$) from the unlabeled set, and allocate accurate labels to them. It is also noted that a few OOD samples (such as the `bear' image in \cref{Fig:ET} being predicted into `dog' with overconfidence) are mixed, but it will be corrected at next epochs. To sum up, the proposed ET dramatically improves the reliability of the model in OOD detection tasks, and finally makes our method converge.

\begin{table*}[btp!]
\setlength{\tabcolsep}{6pt}
\caption{\textbf{Detailed results on CIFAR-10 benchmark using ResNet-18.}
Our method obtains consistently better results across almost all OOD detection metrics and all datasets. ACC shows the classification accuracy on all the ID test samples from $T^I$. $\uparrow$/$\downarrow$ indicates higher/lower value is better.}
\label{T:appendix_cifar10_res18}
\centering 
\begin{tabular}{c|c|ccc|cccc|c}
\toprule
\multirow{2}{*}{Method} & \multirow{2}{*}{Dataset} 
& \multirow{2}{*}{FPR95~$\downarrow$} 
& \multirow{2}{*}{AUROC~$\uparrow$} 
& \multirow{2}{*}{AUPR(In/Out)~$\uparrow$}
& \multicolumn{4}{c|}{CCR@FPR~$\uparrow$} 
& \multirow{2}{*}{ACC~$\uparrow$} \\ 
\cmidrule(lr){6-9}
& && && $10^{-4}$  & $10^{-3}$  & $10^{-2}$ & $10^{-1}$ \\ 
\midrule
\multirow{7}{*}{\begin{tabular}[c]{@{}c@{}}ODIN\end{tabular}}

&Texture     & 42.52          & 84.06          & 86.01          ~/~ 80.73          & 0.02          & 0.18          & 3.71          & 40.14          & 95.02          \\
&SVHN     & 52.27          & 83.26          & 63.76          ~/~ 92.60          & 1.01          & 4.00          & 11.82         & 44.85          & 95.02          \\
& CIFAR-100         & 56.34          & 78.40          & 73.21          ~/~ 80.99          & 0.10          & 0.38          & 4.43          & 30.11          & 95.02          \\
& Tiny-ImageNet & 59.09          & 79.69          & 79.34          ~/~ 77.52          & 0.36          & 0.63          & 4.49          & 34.52          & 92.54          \\
& LSUN &  47.85          & 84.56          & 81.56          ~/~ 85.58          & 0.21          & 0.85          & 9.92          & 46.95          & 95.02          \\
& Places365 &  53.94          & 82.01          & 54.92          ~/~ 93.30          & 0.47          & 1.68          & 7.13          & 39.63          & 93.87          \\
\cmidrule{2-10}
& \textbf{Mean} & \textbf{52.00} & \textbf{82.00} & \textbf{73.13} ~/~ \textbf{85.12} & \textbf{0.36} & \textbf{1.29} & \textbf{6.92} & \textbf{39.37} & \textbf{94.42} \\

\midrule
\multirow{7}{*}{\begin{tabular}[c]{@{}c@{}}EBO\end{tabular}}

&Texture     & 52.11          & 80.70          & 83.34          ~/~ 75.20          & 0.01          & 0.13          & 2.79          & 31.96          & 95.02          \\
&SVHN     & 30.56          & 92.08          & 80.95          ~/~ 96.28          & 1.85          & 5.74          & 21.44         & 75.81          & 95.02          \\
& CIFAR-100         & 56.98          & 79.65          & 75.09          ~/~ 81.23          & 0.10          & 0.69          & 4.74          & 34.28          & 95.02          \\
& Tiny-ImageNet & 57.81          & 81.65          & 81.80          ~/~ 78.75          & 0.33          & 0.95          & 6.01          & 40.40          & 92.54          \\
& LSUN &  50.56          & 85.04          & 82.80          ~/~ 85.29          & 0.24          & 1.96          & 11.35         & 50.43          & 95.02          \\
& Places365 &  52.16          & 83.86          & 58.96          ~/~ 93.90          & 0.39          & 2.11          & 8.38          & 46.00          & 93.87          \\
\cmidrule{2-10}
& \textbf{Mean} & \textbf{50.03} & \textbf{83.83} & \textbf{77.15} ~/~ \textbf{85.11} & \textbf{0.49} & \textbf{1.93} & \textbf{9.12} & \textbf{46.48} & \textbf{94.42} \\

\midrule
\multirow{7}{*}{\begin{tabular}[c]{@{}c@{}}MCD\end{tabular}}

&Texture     & 83.92          & 81.59          & 90.20           ~/~ 63.27          & 4.97          & 10.51         & 29.52          & 62.10           & 90.56          \\
&SVHN     & 60.27          & 89.78          & 85.33          ~/~ 94.25          & 20.05         & 38.23         & 55.43          & 74.01          & 90.56          \\
& CIFAR-100         & 74.00             & 82.78          & 83.97          ~/~ 79.16          & 0.80           & 4.99          & 18.88          & 58.18          & 90.56          \\
& Tiny-ImageNet & 78.89          & 80.98          & 85.63          ~/~ 72.48          & 1.62          & 4.15          & 19.37          & 56.08          & 87.33          \\
& LSUN &  68.96          & 84.71          & 85.74          ~/~ 81.50           & 1.75          & 7.93          & 21.88          & 61.54          & 90.56          \\
& Places365 &  72.08          & 83.51          & 69.44          ~/~ 92.52          & 3.29          & 7.97          & 23.07          & 60.22          & 88.51          \\
\cmidrule{2-10}
& \textbf{Mean} & \textbf{73.02} & \textbf{83.89} & \textbf{83.39} ~/~ \textbf{80.53} & \textbf{5.41} & \textbf{12.30} & \textbf{28.02} & \textbf{62.02} & \textbf{89.68} \\

\midrule
\multirow{7}{*}{\begin{tabular}[c]{@{}c@{}}OE\end{tabular}}

&Texture     & 51.17          & 89.56          & 93.79          ~/~ 81.88          & 6.58           & 11.80           & 27.99          & 71.13         & 91.87          \\
&SVHN     & 20.88          & 96.43          & 93.62          ~/~ 98.32          & 32.72          & 47.33          & 67.20           & 86.75         & 91.87          \\
& CIFAR-100         & 58.54          & 86.22          & 86.17          ~/~ 84.88          & 3.64           & 6.55           & 19.04          & 61.11         & 91.87          \\
& Tiny-ImageNet & 58.98          & 87.65          & 90.9           ~/~ 82.16          & 14.37          & 18.84          & 33.65          & 66.03         & 89.27          \\
& LSUN &  57.97          & 86.75          & 87.69          ~/~ 85.07          & 11.8           & 19.62          & 29.22          & 61.95         & 91.87          \\
& Places365 &  55.64          & 87.00             & 73.11          ~/~ 94.67          & 11.36          & 17.36          & 26.33          & 62.23         & 90.99          \\
\cmidrule{2-10}
& \textbf{Mean} & \textbf{50.53} & \textbf{88.93} & \textbf{87.55} ~/~ \textbf{87.83} & \textbf{13.41} & \textbf{20.25} & \textbf{33.91} & \textbf{68.20} & \textbf{91.29} \\

\midrule
\multirow{7}{*}{\begin{tabular}[c]{@{}c@{}}UDG\end{tabular}}

&Texture     & 20.43          & 96.44          & 98.12          ~/~ 92.91          & 19.90           & 43.33          & 69.19          & 87.71          & 92.94          \\
&SVHN     & 13.26          & 97.49          & 95.66          ~/~ 98.69          & 36.64          & 56.81          & 76.77          & 89.54          & 92.94          \\
& CIFAR-100         & 47.20           & 90.98          & 91.74          ~/~ 89.36          & 1.50            & 10.94          & 40.34          & 75.89          & 92.94          \\
& Tiny-ImageNet & 50.18          & 91.91          & 94.43          ~/~ 86.99          & 0.32           & 23.15          & 53.96          & 78.36          & 90.22          \\
& LSUN &  42.05          & 93.21          & 94.53          ~/~ 91.03          & 14.26          & 37.59          & 60.62          & 81.69          & 92.94          \\
& Places365 &  44.22          & 92.64          & 87.17          ~/~ 96.66          & 10.62          & 35.05          & 58.96          & 79.63          & 91.68          \\
\cmidrule{2-10}
& \textbf{Mean} & \textbf{36.22} & \textbf{93.78} & \textbf{93.61} ~/~ \textbf{92.61} & \textbf{13.87} & \textbf{34.48} & \textbf{59.97} & \textbf{82.14} & \textbf{92.28} \\

\midrule
\multirow{7}{*}{\begin{tabular}[c]{@{}c@{}}Ours\end{tabular}}

&Texture     & 0.78          & 98.92          & 99.55          ~/~ 97.73          & 49.52          & 72.02         & 87.65          & 91.10          & 94.66          \\
&SVHN     & 0.26          & 99.03          & 98.79          ~/~ 99.91         & 64.87         & 82.96         & 91.10          & 93.44          & 94.66          \\
& CIFAR-100         & 29.17          & 91.17          & 92.04          ~/~ 90.02          & 3.16          & 14.07         & 34.89          & 72.02          & 94.66         \\
& Tiny-ImageNet & 7.15          & 94.15          & 97.17          ~/~ 88.57          & 64.27          & 78.25          & 81.40          & 84.20          & 90.41          \\
& LSUN &  0.53          & 98.93          & 99.28          ~/~ 98.91          & 34.88          & 77.74          & 87.65          & 91.10          & 94.66          \\
& Places365 &  13.26          & 96.61          & 95.71          ~/~ 98.78          & 25.16          & 58.64          & 81.40          & 85.79          & 93.23          \\
\cmidrule{2-10}
& \textbf{Mean} & \textbf{8.53} & \textbf{96.47} & \textbf{97.10}~/~\textbf{95.65}  &  \textbf{40.31} & \textbf{63.95} & \textbf{77.35} & \textbf{86.27}  & \textbf{93.71} \\

\bottomrule
\end{tabular}
\end{table*}

\begin{table*}[btp!]
\setlength{\tabcolsep}{6pt}
\caption{\textbf{Detailed results on CIFAR-10 benchmark using WideResNet-28.}
Our method obtains consistently better results across almost all OOD detection metrics and all datasets. ACC shows the classification accuracy on all the ID test samples from $T^I$. $\uparrow$/$\downarrow$ indicates higher/lower value is better.}
\label{T:appendix_cifar10_wrn}
\centering
\begin{tabular}{c|c|ccc|cccc|c}
\toprule
\multirow{2}{*}{Method} & \multirow{2}{*}{Dataset} 
& \multirow{2}{*}{FPR95~$\downarrow$} 
& \multirow{2}{*}{AUROC~$\uparrow$} 
& \multirow{2}{*}{AUPR(In/Out)~$\uparrow$}
& \multicolumn{4}{c|}{CCR@FPR~$\uparrow$} 
& \multirow{2}{*}{ACC~$\uparrow$} \\ 
\cmidrule(lr){6-9}
& && && $10^{-4}$  & $10^{-3}$  & $10^{-2}$ & $10^{-1}$ \\ 
\midrule
\multirow{7}{*}{\begin{tabular}[c]{@{}c@{}}ODIN\end{tabular}}
&Texture     & 47.50          & 81.23          & 82.94          ~/~ 78.25          & 0.00          & 0.00          & 1.81          & 32.69          & 96.08          \\
&SVHN     & 51.17          & 85.36          & 68.02          ~/~ 93.53          & 1.10          & 3.54          & 13.08         & 53.04          & 96.08          \\
& CIFAR-100         & 52.92          & 79.47          & 73.57          ~/~ 82.59          & 0.00          & 0.36          & 3.97          & 30.55          & 96.08          \\
& Tiny-ImageNet & 54.86          & 80.39          & 78.82          ~/~ 79.48          & 0.01          & 0.36          & 3.12          & 33.69          & 93.69          \\
& LSUN &  46.53          & 81.86          & 75.70          ~/~ 85.03          & 0.25          & 0.68          & 3.91          & 33.49          & 96.08          \\
& Places365 &  49.03          & 81.49          & 49.84          ~/~ 93.60          & 0.04          & 0.55          & 3.72          & 33.14          & 95.02          \\
\cmidrule{2-10}
& \textbf{Mean} & \textbf{50.33} & \textbf{81.63} & \textbf{71.48} ~/~ \textbf{85.41} & \textbf{0.23} & \textbf{0.91} & \textbf{4.94} & \textbf{36.10} & \textbf{95.51} \\
\midrule
\multirow{7}{*}{\begin{tabular}[c]{@{}c@{}}EBO\end{tabular}}
&Texture     & 40.44          & 89.55          & 91.16          ~/~ 84.41          & 0.00          & 0.00          & 5.41           & 71.35          & 96.08          \\
&SVHN     & 16.13          & 96.90          & 93.77          ~/~ 98.47          & 2.93          & 18.26         & 68.48          & 91.28          & 96.08          \\
& CIFAR-100         & 42.41          & 88.97          & 85.73          ~/~ 89.42          & 0.01          & 0.72          & 8.77           & 67.94          & 96.08          \\
& Tiny-ImageNet & 45.81          & 89.55          & 89.55          ~/~ 86.72          & 0.03          & 0.61          & 9.93           & 73.79          & 93.69          \\
& LSUN &  37.14          & 90.58          & 87.47          ~/~ 91.07          & 0.29          & 0.83          & 8.51           & 76.21          & 96.08          \\
& Places365 &  39.84          & 89.86          & 68.32          ~/~ 96.33          & 0.04          & 0.68          & 7.15           & 73.24          & 95.02          \\
\cmidrule{2-10}
& \textbf{Mean} & \textbf{36.96} & \textbf{90.90} & \textbf{86.00} ~/~ \textbf{91.07} & \textbf{0.55} & \textbf{3.52} & \textbf{18.04} & \textbf{75.64} & \textbf{95.51} \\
\midrule
\multirow{7}{*}{\begin{tabular}[c]{@{}c@{}}MCD\end{tabular}}
&Texture     & 93.19          & 70.58          & 82.49          ~/~ 49.12          & 0.00             & 0.15          & 7.65           & 44.96         & 87.85          \\
&SVHN     & 88.68          & 81.37          & 74.43          ~/~ 86.75          & 3.28          & 8.65          & 28.28          & 66.86         & 87.85          \\
& CIFAR-100         & 83.29          & 76.58          & 77.17          ~/~ 72.50           & 0.03          & 0.72          & 10.47          & 45.36         & 87.85          \\
& Tiny-ImageNet & 86.6           & 74.83          & 80.53          ~/~ 64.30           & 0.04          & 2.48          & 12.88          & 44.47         & 85.58          \\
& LSUN &  93.06          & 70.14          & 72.62          ~/~ 63.38          & 0.55          & 2.81          & 10.51          & 36.16         & 87.85          \\
& Places365 &  93.13          & 70.42          & 49.04          ~/~ 84.32          & 0.10           & 2.39          & 9.65           & 36.37         & 86.48          \\
\cmidrule{2-10}
& \textbf{Mean} & \textbf{89.66} & \textbf{73.99} & \textbf{72.71} ~/~ \textbf{70.06} & \textbf{0.67} & \textbf{2.87} & \textbf{13.24} & \textbf{45.7} & \textbf{87.24} \\
\midrule
\multirow{7}{*}{\begin{tabular}[c]{@{}c@{}}OE\end{tabular}}
&Texture     & 35.14          & 92.44          & 95.27          ~/~ 87.17          & 5.27          & 8.94           & 31.17          & 79.23          & 94.95          \\
&SVHN     & 22.94          & 96.23          & 94.14          ~/~ 97.78          & 37.34         & 52.79          & 73.87          & 88.74          & 94.95          \\
& CIFAR-100         & 52.99          & 87.17          & 86.80          ~/~ 86.09          & 1.72          & 6.83           & 21.22          & 63.16          & 94.95          \\
& Tiny-ImageNet & 55.53          & 87.43          & 90.20          ~/~ 82.58          & 4.58          & 13.91          & 28.61          & 64.92          & 92.72          \\
& LSUN &  59.69          & 85.56          & 86.18          ~/~ 83.67          & 5.18          & 11.55          & 26.09          & 58.88          & 94.95          \\
& Places365 &  55.30          & 85.75          & 69.15          ~/~ 94.25          & 4.50          & 10.31          & 22.42          & 56.79          & 94.24          \\
\cmidrule{2-10}
& \textbf{Mean} & \textbf{46.93} & \textbf{89.10} & \textbf{86.96} ~/~ \textbf{88.59} & \textbf{9.76} & \textbf{17.39} & \textbf{33.90} & \textbf{68.62} & \textbf{94.46} \\
\midrule
\multirow{7}{*}{\begin{tabular}[c]{@{}c@{}}UDG\end{tabular}}
&Texture     & 22.59          & 95.86          & 97.49          ~/~ 92.59          & 0.87           & 8.92           & 58.06          & 87.56          & 94.50          \\
&SVHN     & 17.23          & 97.23          & 95.43          ~/~ 98.64          & 45.32          & 60.75          & 78.46          & 89.84          & 94.50          \\
& CIFAR-100         & 43.36          & 91.53          & 92.08          ~/~ 90.21          & 5.19           & 12.28          & 37.79          & 77.03          & 94.50          \\
& Tiny-ImageNet & 39.33          & 93.90           & 95.90           ~/~ 90.01          & 4.86           & 27.52          & 64.17          & 82.97          & 92.07         \\
& LSUN &  30.17          & 95.25          & 96.06          ~/~ 94.05          & 13.28          & 36.98          & 66.03          & 86.35          & 94.50          \\
& Places365 &  35.24          & 94.31          & 89.24          ~/~ 97.55          & 8.39           & 27.67          & 61.10           & 83.75          & 93.33         \\
\cmidrule{2-10}
& \textbf{Mean} & \textbf{31.32} & \textbf{94.68} & \textbf{94.36} ~/~ \textbf{93.84} & \textbf{12.98} & \textbf{29.02} & \textbf{60.93} & \textbf{84.58} & \textbf{93.90} \\
\midrule
\multirow{7}{*}{\begin{tabular}[c]{@{}c@{}}Ours\end{tabular}}

&Texture     & 2.03 &99.43 & 99.65          ~/~ 99.02          & 21.81          & 71.75         & 88.68          & 95.07          & 95.73          \\
&SVHN     & 1.13         & 99.87         & 99.72          ~/~ 99.93         & 80.10         & 85.83         & 94.53& 95.61          & 95.37          \\
& CIFAR-100         & 31.40          & 91.43          & 91.03          ~/~ 90.83          & 9.95         & 15.73         & 24.48          & 77.81          & 95.73         \\
& Tiny-ImageNet & 9.37          & 97.18          & 98.35          ~/~ 94.21 & 72.74          & 80.91          & 85.43          & 89.28          & 92.16          \\
& LSUN &  5.18          & 98.83          & 98.92          ~/~ 98.77          & 48.62          & 54.92          & 82.67          & 93.73       & 95.73         \\
& Places365 &  12.49          & 97.25          & 94.15          ~/~ 98.92          & 16.15          & 46.68          & 73.43          & 89.70          & 94.21          \\
\cmidrule{2-10}
& \textbf{Mean} & \textbf{8.24} & \textbf{97.33} & \textbf{96.97}~/~\textbf{96.95}  &  \textbf{41.56} & \textbf{59.26} & \textbf{74.87} & \textbf{90.20}  & \textbf{94.88} \\
\bottomrule
\end{tabular}
\end{table*}

\begin{table*}[btp!]
\setlength{\tabcolsep}{6pt}
\caption{\textbf{Detailed results on CIFAR-100 benchmark using ResNet-18.}
Our method obtains consistently better results across almost all OOD detection metrics and all datasets. ACC shows the classification accuracy on all the ID test samples from $T^I$. $\uparrow$/$\downarrow$ indicates higher/lower value is better.}
\label{T:appendix_cifar100_res18}
\centering
\begin{tabular}{c|c|ccc|cccc|c}
\toprule
\multirow{2}{*}{Method} & \multirow{2}{*}{Dataset} 
& \multirow{2}{*}{FPR95~$\downarrow$} 
& \multirow{2}{*}{AUROC~$\uparrow$} 
& \multirow{2}{*}{AUPR(In/Out)~$\uparrow$}
& \multicolumn{4}{c|}{CCR@FPR~$\uparrow$} 
& \multirow{2}{*}{ACC~$\uparrow$} \\ 
\cmidrule(lr){6-9}
& && && $10^{-4}$  & $10^{-3}$  & $10^{-2}$ & $10^{-1}$ \\ 
\midrule
\multirow{7}{*}{\begin{tabular}[c]{@{}c@{}}ODIN\end{tabular}}

&Texture     & 79.47          & 77.92          & 86.69          ~/~ 62.97          & 2.66          & 4.66          & 15.09          & 45.82          & 76.65          \\
&SVHN     & 90.33          & 75.59          & 65.25          ~/~ 84.49          & 4.98          & 12.02         & 23.79          & 46.61          & 76.65          \\
& CIFAR-10    & 81.82          & 77.90          & 79.93          ~/~ 73.39          & 0.09          & 3.69          & 15.39          & 47.20          & 76.65          \\
& Tiny-ImageNet & 82.74          & 77.58          & 86.26          ~/~ 61.38          & 0.20          & 3.78          & 15.99          & 45.56          & 69.56          \\
& LSUN &  80.57          & 78.22          & 86.34          ~/~ 63.44          & 1.68          & 5.59          & 17.37          & 45.56          & 76.10          \\
& Places365 &  76.42          & 80.66          & 66.77          ~/~ 89.66          & 1.45          & 4.16          & 18.98          & 49.60          & 77.56          \\
\cmidrule{2-10}
& \textbf{Mean} & \textbf{81.89} & \textbf{77.98} & \textbf{78.54} ~/~ \textbf{72.56} & \textbf{1.84} & \textbf{5.65} & \textbf{17.77} & \textbf{46.73} & \textbf{75.53} \\
\midrule
\multirow{7}{*}{\begin{tabular}[c]{@{}c@{}}EBO\end{tabular}}

& Texture     & 84.29          & 76.32          & 85.87          ~/~ 59.12          & 0.82          & 3.89          & 14.37          & 44.60          & 76.65          \\
&SVHN     & 78.23          & 83.57          & 75.61          ~/~ 90.24          & 9.67          & 17.27         & 33.70          & 57.26          & 76.65          \\
& CIFAR-10    & 81.25          & 78.95          & 80.01          ~/~ 74.44          & 0.05          & 4.63          & 18.03          & 48.67          & 76.65          \\
& Tiny-ImageNet & 83.32          & 78.34          & 87.08          ~/~ 62.13          & 1.04          & 6.37          & 21.44          & 47.92          & 69.56          \\
& LSUN &  84.51          & 77.66          & 86.42          ~/~ 61.40          & 1.59          & 6.44          & 19.58          & 46.66          & 76.10          \\
& Places365 &  78.37          & 80.99          & 68.22          ~/~ 89.60          & 1.40          & 4.94          & 21.32          & 51.21          & 77.56          \\
\cmidrule{2-10}
& \textbf{Mean} & \textbf{81.66} & \textbf{79.31} & \textbf{80.54} ~/~ \textbf{72.82} & \textbf{2.43} & \textbf{7.26} & \textbf{21.41} & \textbf{49.39} & \textbf{75.53} \\
\midrule  
\multirow{7}{*}{\begin{tabular}[c]{@{}c@{}}MCD\end{tabular}}
&Texture     & 83.97          & 73.46          & 83.11          ~/~ 56.79          & 0.07          & 1.03          & 9.29           & 38.09          & 68.80          \\
&SVHN     & 85.82          & 76.61          & 65.50          ~/~ 85.52          & 3.03          & 8.66          & 23.15          & 45.44          & 68.80          \\
& CIFAR-10    & 87.74          & 73.15          & 76.51          ~/~ 67.24          & 0.35          & 3.26          & 16.18          & 41.41          & 68.80          \\
& Tiny-ImageNet  & 84.46          & 75.32          & 85.11          ~/~ 59.49          & 0.24          & 6.14          & 19.66          & 41.44          & 62.21          \\
& LSUN & 86.08          & 74.05          & 84.21          ~/~ 58.62          & 1.57          & 5.16          & 18.05          & 41.25          & 67.51          \\
& Places365 & 82.74          & 76.30          & 61.15          ~/~ 87.19          & 1.08          & 3.35          & 14.04          & 43.37          & 70.47          \\
\cmidrule{2-10}
& \textbf{Mean} & \textbf{85.14} & \textbf{74.82} & \textbf{75.93} ~/~ \textbf{69.14} & \textbf{1.06} & \textbf{4.60} & \textbf{16.73} & \textbf{41.83} & \textbf{67.77} \\
\midrule
\multirow{7}{*}{\begin{tabular}[c]{@{}c@{}}OE\end{tabular}}
&Texture     & 86.56          & 73.89          & 84.48          ~/~ 54.84          & 0.66          & 2.86          & 12.86          & 41.81          & 70.49          \\
&SVHN     & 68.87          & 84.23          & 75.11          ~/~ 91.41          & 7.33          & 14.07         & 31.53          & 54.62          & 70.49          \\
& CIFAR-10    & 79.72          & 78.92          & 81.95          ~/~ 74.28          & 2.82          & 9.53          & 23.90          & 48.21          & 70.49          \\
& Tiny-ImageNet & 83.41          & 76.99          & 86.36          ~/~ 60.56          & 0.22          & 8.50          & 21.95          & 43.98          & 63.69          \\
& LSUN &  83.53          & 77.10          & 86.28          ~/~ 60.97          & 1.72          & 7.91          & 22.61          & 44.19          & 69.89          \\
& Places365 &  78.24          & 79.62          & 67.13          ~/~ 88.89          & 3.69          & 7.35          & 20.22          & 47.68          & 72.02          \\
\cmidrule{2-10}
& \textbf{Mean} & \textbf{80.06} & \textbf{78.46} & \textbf{80.22} ~/~ \textbf{71.83} & \textbf{2.74} & \textbf{8.37} & \textbf{22.18} & \textbf{46.75} & \textbf{69.51} \\
\midrule
\multirow{7}{*}{\begin{tabular}[c]{@{}c@{}}UDG\end{tabular}}
&Texture     & 75.04          & 79.53          & 87.63          ~/~ 65.49          & 1.97          & 4.36          & 9.49           & 33.84          & 68.51          \\
&SVHN     & 60.00          & 88.25          & 81.46          ~/~ 93.63          & 14.90         & 25.50         & 38.79          & 56.46          & 68.51          \\
& CIFAR-10    & 83.35          & 76.18          & 78.92          ~/~ 71.15          & 1.99          & 5.58          & 17.27          & 42.11          & 68.51          \\
& Tiny-ImageNet & 81.73          & 77.18          & 86.00          ~/~ 61.67          & 0.67          & 4.82          & 17.80          & 41.72          & 61.80          \\
& LSUN &  78.70          & 76.79          & 84.74          ~/~ 63.05          & 1.59          & 5.34          & 18.04          & 44.70          & 67.10          \\
& Places365 &  73.86          & 79.87          & 65.36          ~/~ 89.60          & 1.96          & 6.33          & 22.03          & 47.97          & 69.83          \\
\cmidrule{2-10}
& \textbf{Mean} & \textbf{75.45} & \textbf{79.63} & \textbf{80.69} ~/~ \textbf{74.10} & \textbf{3.85} & \textbf{8.66} & \textbf{20.57} & \textbf{44.47} & \textbf{67.38} \\
\midrule
\multirow{7}{*}{\begin{tabular}[c]{@{}c@{}}Ours\end{tabular}}

&Texture     & 47.85          & 82.91          & 90.14          ~/~ 67.27         & 0.56          & 2.02         & 23.81          & 52.10          & 73.06          \\
&SVHN     & 7.10          & 95.43          & 93.83          ~/~ 98.31         & 25.22        & 47.87         & 63.36          & 68.39         & 73.06          \\
& CIFAR-10         & 79.25          & 68.20          & 71.40          ~/~ 63.78          & 0.31          & 3.14         & 10.75          & 34.04          & 73.06          \\
& Tiny-ImageNet & 1.95          & 90.28          & 95.23          ~/~ 77.52          & 36.64          & 44.98          & 55.52          & 59.88         & 61.31         \\
& LSUN &  54.09          & 78.34          & 87.37          ~/~ 62.22          & 4.53          & 12.77          & 26.58          & 44.76          & 70.69          \\
& Places365 &  56.08          & 79.45          & 68.25          ~/~ 89.70         & 0.60          & 6.04          & 24.25          & 44.62         & 72.92          \\
\cmidrule{2-10}
& \textbf{Mean}  & \textbf{41.05} & \textbf{82.44}  & \textbf{84.37}~/~\textbf{76.47}  &  \textbf{11.70}  & \textbf{19.47} &  \textbf{34.05} &  \textbf{50.97}  & \textbf{70.70} \\
\bottomrule
\end{tabular}
\end{table*}

\begin{table*}[btp!]
\setlength{\tabcolsep}{6pt}
\caption{\textbf{Detailed results on CIFAR-100 benchmark using WideResNet-28.}
Our method obtains consistently better results across almost all OOD detection metrics and all datasets. ACC shows the classification accuracy on all the ID test samples from $T^I$. $\uparrow$/$\downarrow$ indicates higher/lower value is better.}
\label{T:appendix_cifar100_wrn}
\centering
\begin{tabular}{c|c|ccc|cccc|c}
\toprule
\multirow{2}{*}{Method} & \multirow{2}{*}{Dataset} 
& \multirow{2}{*}{FPR95~$\downarrow$} 
& \multirow{2}{*}{AUROC~$\uparrow$} 
& \multirow{2}{*}{AUPR(In/Out)~$\uparrow$}
& \multicolumn{4}{c|}{CCR@FPR~$\uparrow$} 
& \multirow{2}{*}{ACC~$\uparrow$} \\ 
\cmidrule(lr){6-9}
& && && $10^{-4}$  & $10^{-3}$  & $10^{-2}$ & $10^{-1}$ \\ 
\midrule
\multirow{7}{*}{\begin{tabular}[c]{@{}c@{}}ODIN\end{tabular}}
&Texture     & 78.88          & 76.46          & 84.68          ~/~ 62.45          & 0.15          & 1.52          & 10.21          & 41.44          & 80.25          \\
&SVHN     & 92.26          & 68.41          & 49.07          ~/~ 81.28          & 1.73          & 2.93          & 8.02           & 28.93          & 80.25          \\
& CIFAR-10    & 78.22          & 80.14          & 81.43          ~/~ 76.26          & 0.06          & 3.09          & 15.78          & 50.75          & 80.25          \\
& Tiny-ImageNet & 80.54          & 77.88          & 85.89          ~/~ 62.67          & 0.24          & 2.25          & 13.97          & 45.53          & 72.92          \\
& LSUN &  78.11          & 78.66          & 85.57          ~/~ 65.68          & 0.19          & 1.26          & 11.69          & 45.32          & 78.54          \\
& Places365 &  73.62          & 80.57          & 63.79          ~/~ 90.13          & 0.86          & 2.79          & 13.03          & 47.47          & 80.03          \\
\cmidrule{2-10}
& \textbf{Mean} & \textbf{80.27} & \textbf{77.02} & \textbf{75.07} ~/~ \textbf{73.08} & \textbf{0.54} & \textbf{2.31} & \textbf{12.12} & \textbf{43.24} & \textbf{78.71} \\
\midrule
\multirow{7}{*}{\begin{tabular}[c]{@{}c@{}}EBO\end{tabular}}
&Texture     & 84.22          & 76.13          & 85.08          ~/~ 58.51          & 0.08          & 1.55          & 10.04          & 44.24          & 80.25          \\
&SVHN     & 80.05          & 79.88          & 65.44          ~/~ 88.37          & 0.97          & 3.88          & 14.93          & 50.85          & 80.25          \\
& CIFAR-10    & 76.18          & 81.50          & 83.34          ~/~ 77.36          & 0.45          & 6.11          & 21.03          & 53.73          & 80.25          \\
& Tiny-ImageNet & 80.78          & 79.94          & 88.02          ~/~ 64.18          & 0.06          & 4.92          & 22.31          & 51.82          & 72.92          \\
& LSUN &  82.59          & 78.74          & 86.71          ~/~ 62.94          & 0.64          & 1.55          & 17.71          & 49.76          & 78.54          \\
& Places365 &  74.54          & 81.63          & 67.67          ~/~ 90.18          & 1.13          & 3.69          & 17.55          & 52.47          & 80.03          \\
\cmidrule{2-10}
& \textbf{Mean} & \textbf{79.73} & \textbf{79.64} & \textbf{79.38} ~/~ \textbf{73.59} & \textbf{0.55} & \textbf{3.62} & \textbf{17.26} & \textbf{50.48} & \textbf{78.71} \\
\midrule
\multirow{7}{*}{\begin{tabular}[c]{@{}c@{}}MCD\end{tabular}}
&Texture     & 91.33          & 69.03          & 79.60          ~/~ 49.66          & 0.00          & 0.29          & 4.49           & 32.61          & 68.80          \\
&SVHN     & 87.03          & 73.48          & 52.89          ~/~ 84.73          & 1.74          & 2.90          & 6.68           & 33.88          & 68.80          \\
& CIFAR-10    & 86.89          & 73.79          & 76.15          ~/~ 68.38          & 0.26          & 2.88          & 13.40          & 39.94          & 68.80          \\
& Tiny-ImageNet & 85.16          & 74.59          & 84.19          ~/~ 58.36          & 1.01          & 2.58          & 13.71          & 40.31          & 62.22          \\
& LSUN &  88.67          & 72.04          & 83.06          ~/~ 54.33          & 1.13          & 3.58          & 15.95          & 39.58          & 67.29          \\
& Places365 &  86.83          & 74.05          & 59.58          ~/~ 85.28          & 1.24          & 3.66          & 14.85          & 41.07          & 69.77          \\
\cmidrule{2-10}
& \textbf{Mean} & \textbf{87.65} & \textbf{72.83} & \textbf{72.58} ~/~ \textbf{66.79} & \textbf{0.90} & \textbf{2.65} & \textbf{11.51} & \textbf{37.90} & \textbf{67.61} \\
\midrule
\multirow{7}{*}{\begin{tabular}[c]{@{}c@{}}OE\end{tabular}}
&Texture     & 93.07          & 67.00             & 78.92     ~/~ 46.52          & 0.02         & 0.52          & 5.50            & 32.16         & 74.01         \\
&SVHN     & 88.74          & 76.14          & 66.07          ~/~ 85.17          & 7.06         & 12.91         & 24.82          & 47.43         & 74.01         \\
& CIFAR-10    & 78.82          & 79.36          & 81.29          ~/~ 75.27          & 1.08         & 7.63          & 17.49          & 48.84         & 74.01         \\
& Tiny-ImageNet & 83.34          & 78.35          & 87.34          ~/~ 61.78          & 1.06         & 8.84          & 24.40           & 47.64         & 66.49         \\
& LSUN &  84.96          & 78.11          & 87.26          ~/~ 60.76          & 5.80          & 10.40          & 25.75          & 48.27         & 71.47         \\
& Places365 &  80.30           & 79.87          & 67.23          ~/~ 88.65          & 1.78         & 6.29          & 19.78          & 49.84         & 74.39         \\
\cmidrule{2-10}
& \textbf{Mean} & \textbf{84.87} & \textbf{76.47} & \textbf{78.02} ~/~ \textbf{69.69} & \textbf{2.80} & \textbf{7.76} & \textbf{19.63} & \textbf{45.70} & \textbf{72.40} \\
\midrule
\multirow{7}{*}{\begin{tabular}[c]{@{}c@{}}UDG\end{tabular}}
&Texture     & 73.62          & 79.01          & 85.53          ~/~ 67.08          & 0.00             & 0.00             & 6.74           & 46.09         & 75.77          \\
&SVHN     & 66.76          & 85.29          & 76.14          ~/~ 92.33          & 8.00             & 15.83         & 32.57          & 58.05         & 75.77          \\
& CIFAR-10    & 82.35          & 76.67          & 78.52          ~/~ 72.63          & 0.51          & 3.90           & 15.29          & 44.79         & 75.77          \\
& Tiny-ImageNet & 78.91          & 79.04          & 87.00             ~/~ 65.06          & 0.12          & 2.86          & 19.13          & 47.50          & 68.57          \\
& LSUN &  77.04          & 79.79          & 87.49          ~/~ 66.93          & 2.51          & 6.01          & 22.33          & 49.14         & 73.93          \\
& Places365 &  72.25          & 81.49          & 66.72          ~/~ 90.65          & 1.19          & 3.28          & 17.59          & 50.82         & 76.10           \\
\cmidrule{2-10}
& \textbf{Mean} & \textbf{75.16} & \textbf{80.21} & \textbf{80.23} ~/~ \textbf{75.78} & \textbf{2.05} & \textbf{5.31} & \textbf{18.94} & \textbf{49.40} & \textbf{74.32} \\
\midrule
\multirow{7}{*}{\begin{tabular}[c]{@{}c@{}}Ours\end{tabular}}

&Texture     & 34.47          & 73.50          & 85.15          ~/~ 55.77          & 0.73          & 4.81         & 21.51          & 40.96          & 76.47          \\
&SVHN     & 7.71          & 96.22          & 93.59          ~/~ 97.92         & 19.61         & 36.21         & 58.39          & 71.37          & 76.85          \\
& CIFAR-10        & 56.96          & 66.97          & 72.19          ~/~ 61.96          & 2.40          & 7.66         & 16.52          & 33.95          & 76.82         \\
& Tiny-ImageNet & 2.53          & 89.55          & 94.95          ~/~ 75.10          & 36.59          & 42.33         & 52.94         & 61.37          & 66.12          \\
& LSUN &  72.63          & 69.87          & 81.48          ~/~ 52.22         & 0.67          & 5.25          & 14.52          & 33.03         & 69.54          \\
& Places365 &  49.73         & 79.88          & 67.82          ~/~ 89.04         & 3.03          & 8.35         & 22.07         & 46.90          & 76.80          \\
\cmidrule{2-10}
& \textbf{Mean} & \textbf{38.19} & \textbf{83.01} & \textbf{84.90}~/~\textbf{76.60}  &  \textbf{10.20} & \textbf{18.57} & \textbf{34.17} & \textbf{52.70}  & \textbf{74.89} \\
\bottomrule
\end{tabular}
\end{table*}

\section{Detailed Results and More Architectures}

\cref{T:appendix_cifar10_res18} and \cref{T:appendix_cifar100_res18} show the detailed results among all datasets. Our method obtains consistently better results across all OOD detection metrics and all datasets. Compared with other methods using extra OOD training data (MCD~\cite{mcd}, OE~\cite{hendrycks18oe}, UDG~\cite{yang2021semantically}), our method boosts  the OOD detection performance meanwhile maximally maintaining the ID classification performance  and achieves the best results on ACC.
Following~\cite{yang2021semantically} we also adopt another network architecture of  WideResNet-28~\cite{resnet} to do experiments in \cref{T:appendix_cifar10_wrn} and  \cref{T:appendix_cifar100_wrn} and compare the performance. The comparison results on WideResNet-28 have the same trend as that on ResNet-18~\cite{resnet} architecture. Our proposed method combining ET with $L_{rep}$ has advantages on almost all metrics, showing that our method enhances both the OOD detection and the ID classification ability. 
Notably, the previous state-of-the-art approaches generally performed well on SVHN and Texture datasets, but in Tiny-ImageNet, LSUN, and Places365 suffered a defeat. It can be explained that the images in the first two datasets have relatively flat backgrounds, which are quite different in style from those in CIFAR10/100, and the resulting covariate shifts make it easier for the model to identify them as OOD examples. Our proposed method especially achieves performance advantages on Tiny-ImageNet, LSUN, and Places365, but its performance on CIFAR-100/10 (the styles between them are very similar) of the CIFAR-10/100 benchmark is still not outstanding. This indicates that more 
exploration is needed to overcome the interference of covariate shifts in OOD detection tasks.

\begin{figure*}[!t]
\centering
\includegraphics[width=\linewidth]{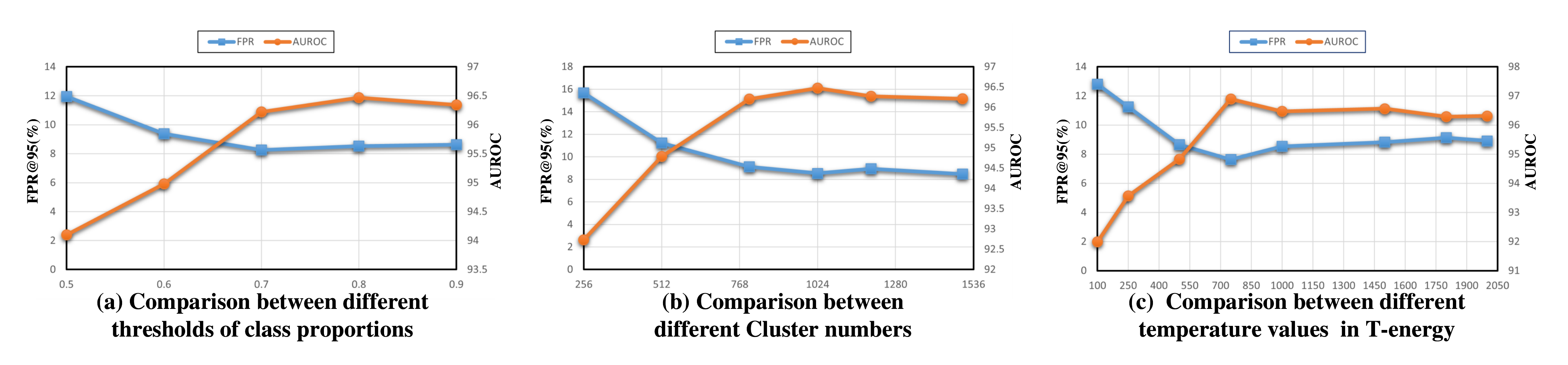}
\vspace{-20pt}
\caption{\textbf{Discuss about hyper-parameters.} (a), (b), and (c) respectively demonstrate the impact of the threshold of class proportion $\tau$, clusters numbers $K$, and temperature value $T$ on the performance of our method. All the performance fluctuation is very small when $\tau\ge0.7$, $K\ge800$ and $T\ge750$, showing the robustness of our method.}
\label{fig:hyper-par}
\end{figure*}

\begin{table*}[btp!]
\renewcommand{\arraystretch}{1.}
\renewcommand{\tabcolsep}{8.pt}
\centering
\small
\caption{\textbf{Comparison between the previous SOTA methods and ours on a large-scale benchmark.} Our method obtains the best results across almost all OOD detection metrics. $\uparrow$/$\downarrow$ indicates higher/lower value is better and the best results are in \textbf{bold}.}
\label{T:large scale}
\begin{tabular}{cc|ccc|cccc}
\toprule
\multirow{2}{*}{ID data} 
& \multirow{2}{*}{Method} 
& \multirow{2}{*}{FPR95~$\downarrow$}  
& \multirow{2}{*}{AUROC~$\uparrow$} 
& \multirow{2}{*}{AUPR-In/Out~$\uparrow$}
& \multicolumn{4}{c}{CCR@FPR~$\uparrow$} \\ 
\cmidrule(lr){6-9}
& && && $10^{-4}$  & $10^{-3}$  & $10^{-2}$ & $10^{-1}$                      \\ \midrule
\multirow{6}{*}{\begin{tabular}[c]{@{}c@{}}100 classes from\\ ImageNet\end{tabular}}
& ODIN~\cite{odin}         &  80.07 & 51.23 & 56.93~/~50.46 & 0.06 &1.18 & 8.42  & 13.36 \\ 
& EBO~\cite{energyood}      &  82.77 &	50.17 &	55.31~/~49.84 & 0.49 &1.49	&8.87 &13.59 \\
& OE~\cite{hendrycks18oe}  & 80.52	& 55.87	& 55.93~/~51.94	& 1.06	& \textbf{2.63}	& 7.67	& 15.13 \\
& MCD~\cite{mcd}           & 91.04  &52.26 &	54.80~/~43.92 &0.08	 &1.92  &5.57 &	14.35  \\
& UDG\cite{yang2021semantically}   & 81.89 & 54.74 & 57.85~/~52.53 & 0.95  & 2.06 & 9.18 &  16.35\\
& \textbf{Ours}     & \textbf{60.17} & \textbf{63.91} & \textbf{69.55}~/~\textbf{58.23}  &  \textbf{1.08} & 2.16 & \textbf{10.56} & \textbf{21.34}          \\ 
\bottomrule
\end{tabular}
\vspace{-0.20em}
\end{table*}

\section{Choice of hyper-parameters.}
Here we analyze the impact of the main hyper-parameters including the threshold of class proportion $\tau$, clusters numbers $K$, and temperature value $T$, and prove the robustness of the proposed method. \cref{fig:hyper-par}\textcolor{red}{-a} shows that the maximum fluctuation is $0.38\%$/$0.25\%$ on FPR@95/AUROC when $\tau$=0.7-0.9,  and \cref{fig:hyper-par}\textcolor{red}{-b} indicates when $K$=800-1200, the maximum fluctuation is $0.58\%$/$0.27\%$ on FPR@95/AUROC. We choose $\tau$=0.8 and $K$=1024 following \cite{yang2021semantically} in the paper for fair comparisons. During rebuttal, we added the experiment shown in \cref{fig:hyper-par}\textcolor{red}{-c} and demonstrate the results are insensitive to large temperature values $T$ (\textit{e.g., }$\ge750$). The best $T$ is around 750 but we also achieve good results when choosing $T$=1000 in the paper.

\section{Experiments on large-scale datasets.} 
To evaluate the generalization of our method in realistic scenarios, we extend it to large-scale datasets. Specially, we choose 50,000 samples from 100 classes in ImageNet as labeled ID training set $D_L$ and still use Tiny-ImageNet as the unlabeled training set $D_U$ mixed with ID and OOD data. The testing set $T$ contains 1,000 ID images and 4,000 OOD images from ImageNet. The comparison results in \cref{T:large scale} demonstrate that our proposed uncertainty-aware optimal transport scheme obtains the best results on almost all metrics, indicating its generalization on large-scale datasets.
However, all methods in \cref{T:large scale} exhibit significant performance degradation on the large-scale datasets, which suggests that more effort is needed to address the challenges presented by benchmarks with more diverse categories and higher resolution in the OOD detection area.

\end{document}